\documentclass[journal]{IEEEtran}  
\IEEEoverridecommandlockouts

\usepackage{xcolor}
\usepackage{graphics} 
\usepackage{epsfig} 
\usepackage{amsmath} 
\usepackage{amssymb}  
\usepackage{enumerate}
\newtheorem{remark}{Remark}
\usepackage{afterpage}
\usepackage{euscript}
\usepackage{multirow}
\usepackage{xparse}
\usepackage{algorithm}
\usepackage[]{algpseudocode}

\usepackage{float}
\newfloat{algorithm}{t}{lop}

\newcommand{\doi}{10.1109/TRO.2020.2964147}

\newcounter{SetupCount}
\newenvironment{changemargin}[2]{%
\begin{list}{}{%
\setlength{\topsep}{0pt}%
\setlength{\leftmargin}{#1}%
\setlength{\rightmargin}{#2}%
\setlength{\listparindent}{\parindent}%
\setlength{\itemindent}{\parindent}%
\setlength{\parsep}{\parskip}%
}%
\item[]}{\end{list}}

\usepackage{mathtools}
\title{ 
\Huge %
Using Human Ratings for Feedback Control:
A Supervised Learning Approach with Application to Rehabilitation Robotics
}
\author{Marcel Menner, Lukas Neuner, Lars L\"unenburger, and Melanie N. Zeilinger
\thanks{
This work was supported by the Swiss National Science Foundation under Grant PP00P2$\_$157601/1.
}
        \thanks{M. Menner, L. Neuner, and M.N. Zeilinger are with the Institute for Dynamic Systems and Control,
       ETH Zurich, 8092 Zurich, Switzerland   (e-mail:
        mmenner@ethz.ch; lneuner@ethz.ch; mzeilinger@ethz.ch).}%
        \thanks{L. L\"unenburger is with Hocoma AG,
       8604 Volketswil, Switzerland   (e-mail: lars.luenenburger@hocoma.com).}%
\thanks{Digital Object Identifier \doi}%
}

\usepackage{textcomp}
\usepackage{bbding}
\newcommand{\tp}{\mathrm{T}}
\usepackage{pifont}
\usepackage{cite}
\usepackage{wasysym}

\begin{document}

\titlepage
\vspace*{3cm}
\begin{center}
\LARGE \bf
Using Human Ratings for Feedback Control:
A Supervised Learning Approach with Application to Rehabilitation Robotics
\end{center}
\vspace{0.0cm}
\begin{center}
Marcel Menner, Lukas Neuner, Lars L\"unenburger, and Melanie N. Zeilinger
\end{center}

\vspace{3cm}
\noindent This work has been accepted for publication in the IEEE Transactions on Robotics.

\noindent Digital Object Identifier \doi.

\vspace{1cm}
\noindent Please cite this work as
\\

\noindent  M. Menner, L. Neuner, L. L\"unenburger, and M. N. Zeilinger, "Using Human Ratings for Feedback Control:
A Supervised Learning Approach with Application to Rehabilitation Robotics," \textit{IEEE Transactions on Robotics}, 2020, doi:\doi.
\vspace*{\fill}

\begin{changemargin}{0cm}{0cm}
\noindent 
\textcopyright\hspace{-0.1cm} 2019 IEEE.  
 Personal use of this material is permitted.  Permission from IEEE must be obtained for all other uses, in any current or future media, including reprinting/republishing this material for advertising or promotional purposes, creating new collective works, for resale or redistribution to servers or lists, or reuse of any copyrighted component of this work in other works.
\end{changemargin}

\twocolumn
\newpage

\maketitle
\thispagestyle{empty}
\pagestyle{empty}

\begin{abstract}
This article presents a method for tailoring a parametric controller based on human ratings. 
The method leverages supervised learning concepts in order to train a reward model from data. 
It is applied to a gait rehabilitation robot with the goal of teaching the robot how to walk patients physiologically.
In this context, the reward model judges the physiology of the gait cycle (instead of therapists) using sensor measurements provided by the robot and the automatic feedback controller chooses the input settings of the robot to maximize the reward.
The key advantage of the proposed method is that only a few input adaptations are necessary to achieve a physiological gait cycle.
Experiments with nondisabled subjects show that the proposed method permits the incorporation of human expertise into a control law and to automatically walk patients physiologically. 
\end{abstract}
\begin{IEEEkeywords}
Human feedback-based control, human-centered robotics, learning and adaptive systems, rehabilitation robotics. 
\end{IEEEkeywords}
\section{Introduction}
\begin{figure}[t]
      \centering
     \includegraphics[width=0.9\columnwidth]{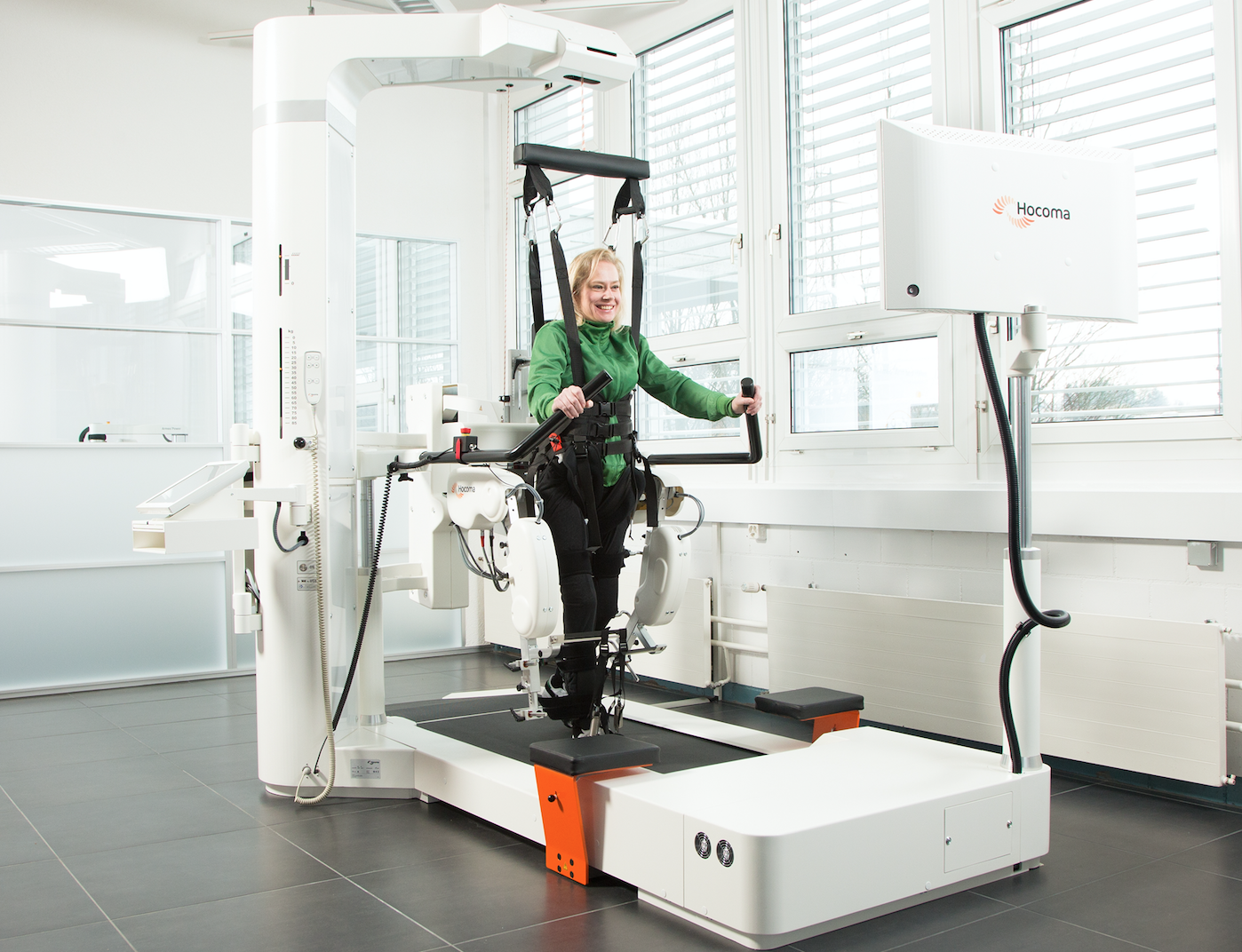}
      \caption{Lokomat\textsuperscript{\textregistered} gait rehabilitation robot (Hocoma AG, Volketswil, CH). 
      }
      \label{fig:lokomat}
\end{figure}

\IEEEPARstart{H}{umans}
can perform very complex tasks that are difficult to achieve with autonomous systems.
The dependency on human supervision or expertise still restricts efficient operation of many complex systems.
An important domain where human expertise is usually needed is rehabilitation robotics, where we consider the robot-assisted gait trainer {Lokomat}\textsuperscript{\textregistered} \cite{colombo2000} in this paper, see Fig.~\ref{fig:lokomat}.
Robotic systems like the Lokomat have recently been introduced in gait rehabilitation following neurological injuries with the goal of mitigating the limitations of conventional therapy \cite{reinkensmeyer2004robotics,emken2005robot,marchal2009review,lambercy2012robots,van2016rehabilitation}. 
However, training with such robots still requires the supervision and interaction of experienced therapists \cite{colombo2000}.

Gait rehabilitation with the Lokomat currently requires physiotherapists to manually adjust the mechanical setup and input settings, e.g., the speed of the treadmill or the range of motion, in order to bring patients into a physiological and safe gait cycle. 
Therapists have to be trained specifically for the device and acquire substantial experience in order to achieve good input settings. 
Although there are guidelines for their adjustment \cite{HocomaNotes15}, it remains a heuristic process, which strongly depends on the knowledge and experience of the therapist.
Automatic adaptation of input settings can reduce the duration of therapists' schooling, improve patient training, make the technology more broadly applicable, and can be more cost effective.
In this work, we propose a method to automatically adapt the input settings.
Although the motivation behind this work is in the domain of rehabilitation robotics, the proposed method addresses general human-in-the-loop scenarios, where expert knowledge can improve system operation.

In this paper, we propose a two-step approach to achieve automatic input adaptations:
First, we define a feature vector to characterize the gait cycle and postulate a reward model to judge the physiology of the gait cycle using the feature vector.
The reward model is trained with therapists' ratings using a supervised learning technique, where the feature vector is obtained from sensor measurements provided by the robot.
The sensor measurements are the angle, torque, and power of both the hip and knee joints of the robot.
Second, we use the gradient of the reward model to determine input adaptations that achieve the desired gait cycle.
This involves a steady-state model to relate the gradient of the reward model with respect to the feature vector (high dimensional) to input settings (low dimensional) that adjust the gait cycle.
A key component in the proposed formulation is that the reward model and its gradient are formulated as functions of the feature vector rather than the input settings.
The high dimensionality of the feature vector allows us to use one model for all human subjects with very different body types, which enables very efficient online application of the proposed method.
In order to train both the reward model and the steady-state model, we collected data with various physiological and non-physiological input settings from 16 nondisabled subjects. 
The subjects were instructed to be passive while being walked by the robot in order to imitate patients with limited or no ability to walk in the early stages of recovery.
Experiments with ten nondisabled subjects highlighted the ability of the proposed method to improve the walking pattern within few adaptations starting from multiple initially non-physiological gait cycles.

\subsection*{Related Work}
Adaptive control strategies have been the subject of a body of research in robotic gait trainers with the goal of improving the therapeutic outcome of treadmill training \cite{Colombo2004, Riener2005, Riener2006, Duschau2007, duschau2010, Maggioni2015}.
The work in \cite{Colombo2004} presents multiple strategies for automatic gait cycle adaptation in robot-aided gait rehabilitation based on minimizing the interaction torques between device and patient. 
Biomechanical recordings providing feedback about a patient's activity level are introduced in \cite{Riener2005,Riener2006}. 
Automated synchronization between treadmill and orthosis based on iterative learning is introduced in \cite{Duschau2007}. 
In \cite{duschau2010}, a path control method is proposed to allow voluntary movements along a physiological path defined by a virtual tunnel.  
An algorithm to adjust the mechanical impedance of an orthosis joint based on the level of support required by a patient is proposed in \cite{Maggioni2015}. 
Further research in the domain of rehabilitation robotics is presented, e.g., in \cite{emken2007motor,koenig2011psychological}.
In \cite{emken2007motor}, the human motor system is modeled and analyzed as approximating an optimization problem trading off effort and kinematic error.  
In \cite{koenig2011psychological}, a patient's psychological state is estimated to judge their mental engagement.
Different from the work in rehabilitation robotics presented in  \cite{Colombo2004,Riener2005,Riener2006,Duschau2007,duschau2010, Maggioni2015, emken2007motor, koenig2011psychological}, we present a method for input setting adaptation of a rehabilitation robot based on a feedback controller, which is derived from human ratings.

In the following, we discuss research directions related to the techniques employed in the proposed approach. 

\textit{Gait cycle classification} is often used to distinguish human subjects according to two classes \cite{Begg2005,Wu2007,Yang2012,Tahafchi2017}, e.g., young/elderly or healthy/impaired.
In \cite{Begg2005}, a supervised learning method for automatic recognition of movement patterns is presented to discriminate gait patterns of young and elderly people.
In order to improve classification performance, \cite{Wu2007} employs a kernel-based principle component analysis for the extraction of features.
Gait patterns are also used to diagnose diseases that symptomatically cause gait abnormalities, e.g.,
 \cite{Yang2012,Tahafchi2017}.
Different from \cite{Begg2005,Wu2007,Yang2012,Tahafchi2017}, this paper does not aim to identify or classify human individuals but to generalize from data of multiple individuals by classifying gait patterns according to their physiology.
Further, the obtained classifier is not used to predict discrete/binary classes but as a continuous reward, which is maximized using feedback control.

\textit{Reinforcement learning}  uses a trial and error search to find a control policy \cite{Sutton1998,Knox2009, Pilarski2011, Griffith2013, Christiano2017, wilde2018learning, dorsa2017active, basu2018learning, Menner2018}.
The framework proposed in \cite{Knox2009} allows human trainers to shape a policy using approval or disapproval.  
In \cite{Pilarski2011}, human-generated rewards in a reinforcement learning framework are employed for a 2-joint velocity control task. 
In \cite{Griffith2013}, a policy shaping method is presented where human feedback is not used as a reward signal but directly as a label for the policy.  
In \cite{Christiano2017}, human preferences are learned through ratings based on a pairwise comparison of trajectories with the goal of reducing human feedback.
In \cite{wilde2018learning}, a robot motion planning problem is considered, where users provide a ranking of paths that enable the evaluation of the importance of different constraints. 
In \cite{dorsa2017active}, a method is presented that actively synthesizes queries to a user to update a distribution over reward parameters.
In \cite{basu2018learning}, user preferences in a traffic scenario are learned based on human guidance by means of feature queries.
In \cite{Menner2018}, human ratings are used to learn a probability distribution of individual preferences modeled as a Markov decision process.
While a reinforcement learning approach could in principle be applied to the considered problem, the online application of these methods typically requires a few hundred human ratings to learn a policy. 
This is infeasible when working with a patient, where a comparatively small number of feedback rounds has to be sufficient. 
The main difference between our method and reinforcement learning is that we do not use trial and error search but we build a reward model that is maximized online.

\textit{Inverse optimal control} and \textit{inverse reinforcement learning} aim at learning a reward model or cost model from demonstrations of human behavior \cite{Kalman1964,ng2000algorithms, hewing2019learning, ravichandar2020recent, Mombaur2010human,clever2016inverse, kleesattel2018inverse, menner2018predictive, lin2016human, Englert2017, Menner2019cdc, Menner2018ecc,abbeel2004apprenticeship, ziebart2008maximum, levine2011nonlinear, Hadfield2016, bogert2016expectation, joukov2017gaussian}. 
Inverse optimal control methods model demonstrations to be the result of an optimal control problem \cite{Mombaur2010human, clever2016inverse, lin2016human, Englert2017, Menner2019cdc,Menner2018ecc, menner2018predictive, kleesattel2018inverse} and often aim at transferring human expertise to an autonomous system, e.g., for humanoid locomotion \cite{Mombaur2010human,clever2016inverse}, identifying human movements \cite{lin2016human, kleesattel2018inverse, menner2018predictive}, robot manipulation tasks \cite{Englert2017}, or autonomous driving \cite{Menner2019cdc}.
In inverse reinforcement learning \cite{abbeel2004apprenticeship, ziebart2008maximum, levine2011nonlinear, Hadfield2016, bogert2016expectation,joukov2017gaussian}, demonstrations are typically modeled to be the result of probabilistic decision-making in a Markov decision process. 
The fundamental difference between the proposed method and inverse optimal control/inverse reinforcement learning methods is the utilization of ratings instead of demonstrations to learn a reward model.

\section{Hardware Description \& Problem Definition} 
The Lokomat\textsuperscript{\textregistered} gait rehabilitation robot (Hocoma AG, Volketswil, CH) is a bilaterally driven gait orthosis that is attached to the patient's legs by Velcro straps.
In conjunction with a bodyweight support system, it provides controlled flexion and extension movements of the hip and knee joints in the sagittal plane. 
Leg motions are repeated based on predefined but adjustable reference trajectories. 
Additional passive foot lifters ensure ankle dorsiflexion during swing. 
The bodyweight support system partially relieves patients from their bodyweight via an attached harness. 
A user interface enables gait cycle adjustments by therapists via a number of input settings \cite{colombo2000,Riener2006}. 
 
\subsubsection*{Input Settings}
One important task of the therapist operating the Lokomat is the adjustment of the input settings to obtain a desirable gait trajectory.
A total of 13 input settings can be adjusted to affect the walking behavior, which are introduced in Table~\ref{tb:settings}.
In this work, we propose a method that can automate or assist the therapists in the adjustment of the input settings by measuring the gait cycle.

\begin{table}[t]
\caption{
Input Settings of the Lokomat
}
\begin{center}
\def\arraystretch{1.00}
\setlength{\tabcolsep}{3pt}
\label{tb:settings}
\begin{tabular}{lcc }
\hline
Input Setting \& Description & Step-size & Range  
\\
\hline
\vspace{-0.25cm}
\\
\textbf{Hip Range of Motion} (Left \& Right) 
& 3${}^\circ$ & 23${}^\circ$, 59${}^\circ$ 
\\
\multicolumn{3}{l}{Defines the amount of flexion and extension}
\vspace{0.05cm}
\\
\textbf{Hip Offset} (Left \& Right)
& 1${}^\circ$  & -5${}^\circ$, 10${}^\circ$
\\
\multicolumn{3}{l}{Shifts movements towards extension or flexion}
\vspace{0.05cm}
\\
\textbf{Knee Range of Motion} (Left \& Right)
& 3${}^\circ$  & 32${}^\circ$, 77${}^\circ$
\\
\multicolumn{3}{l}{Defines amount of flexion}
\vspace{0.05cm}
\\
\textbf{Knee Offset} (Left \& Right)
& 1${}^\circ$   & 0${}^\circ$, 8${}^\circ$
\\
\multicolumn{3}{l}{Shifts movement into flexion for hyperextension correction}
\vspace{0.05cm}
\\
\textbf{Speed}
& 0.1km/h  & 0.5km/h, 3km/h  
\\
\multicolumn{3}{l}{Sets the treadmill speed}
\vspace{0.05cm}
 \\
\textbf{Orthosis speed } 
& 0.01 & 0.15, 0.8
\\
\multicolumn{3}{l}{Defines the orthosis and affects walking cadence}
\vspace{0.05cm}
 \\
\textbf{Bodyweight Support}
& continuous & 0kg, 85kg
\\
\multicolumn{3}{l}{Defines carried weight for unloading}
\vspace{0.05cm}
 \\
\textbf{Guidance Force}
& 5\% & 0\%, 100\% 
\\
\multicolumn{3}{l}{Sets amount of assistance}
\vspace{0.05cm}
 \\
\textbf{Pelvic}
& 1cm  & 0cm, 4cm
\\
\multicolumn{3}{l}{Defines lateral movement}
 \\
\hline
\end{tabular}
\end{center}
\end{table}

\subsection{State-of-the-Art Therapy Session}
The current practice of gait rehabilitation with the Lokomat includes the preparation and setup of the patient and device, actual gait training, and finally removing the patient from the system \cite{HocomaNotes15}. 
Gait training is further divided into three phases:
\begin{enumerate}[1.]
\item Safe walk: The patient is gradually lowered until the dynamic range for the bodyweight support is reached. 
The purpose of this first phase is to ensure a safe and non-harmful gait cycle.  
\item Physiological walk: After ensuring safe movements, the gait cycle is adjusted so that the patient is walked physiologically by the robot. 
\item Goal-oriented walk: The gait cycle is adjusted to achieve therapy goals for individual sessions while ensuring that the patient's gait remains physiological. 
\end{enumerate}

In this paper, we focus on the physiological walk.
In a state-of-the-art therapy session, therapists are advised to follow published heuristic guidelines on how to adjust the input settings based on observations (visual feedback) in order to reach a physiological walk.
Three examples of the heuristic guidelines are as follows: 
If the step length does not match walking speed, then the hip range of motion or treadmill speed should be adjusted; 
if the heel strike is too late, then the hip offset or the hip range of motion should be decreased; 
if the foot is slipping, then the orthosis speed or the knee range of motion should be decreased.
An extended overview of heuristics can be found in \cite{HocomaNotes15}.
This heuristic approach requires experience and training with experts, which incurs high costs and limits the availability of the rehabilitation robot due to the small number of experienced experts. 
The proposed method aims to alleviate this limitation as described in the following.

\subsection{Technological Contribution}
\begin{figure}[t]
      \centering
     \includegraphics[width=0.96\columnwidth]{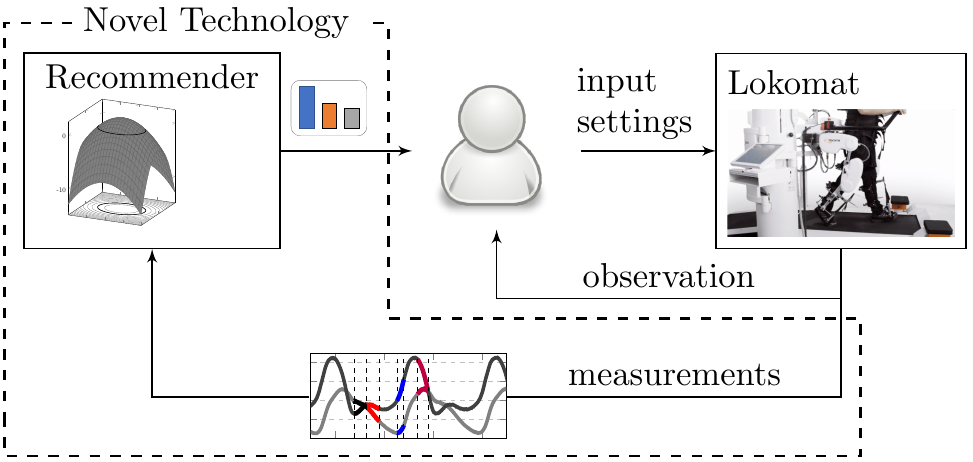}
      \caption{Overview of the proposed method as a recommendation system. 
       The novel technology (dashed lines) augments the state-of-the-art control loop of a therapist and the Lokomat.
Sensor measurements of angle, torque, and power of both hip and knee joints provided by the Lokomat are used to compute recommendations for the input adaptations.  }
      \label{fig:loop}
\end{figure}

We propose a method for automatically suggesting suitable input settings for the Lokomat based on available sensor measurements in order to walk patients physiologically. 
The proposed framework can be used for recommending input settings for therapists,  automatic adaptation of input settings, or as an assistive method for therapists during schooling with the Lokomat. 
Fig.~\ref{fig:loop} illustrates the proposed method as a recommendation system.
The method is derived assuming that the mechanical setup of the Lokomat is done properly, such that the purpose of adapting the input settings is the improvement of the gait cycle and not corrections due to an incorrect setup.
\section{Controller Design based on Human Ratings}
\label{sec:theory}

This section describes the proposed human feedback-based controller.
In the setup considered, input settings $\boldsymbol s\in \mathbb{R}^m$ of the controlled system lead to a gait cycle represented by a feature vector $\boldsymbol x\in \mathbb{R}^n$ in steady-state:
\begin{align}
\label{eq:sys}
\boldsymbol x=\boldsymbol f(\boldsymbol s),
\end{align}
where $\boldsymbol f$ is an unknown function. 
For the considered application, the input settings $\boldsymbol s$ are given in Table~\ref{tb:settings} and the feature vector $\boldsymbol x$ is composed of statistical features of measurements, which characterize the gait cycle and are further discussed in Section~\ref{sec:lokomat}.
Here, the notion of a steady-state means that any transients due to an input adaptation have faded.
The control objective is to find input settings $\boldsymbol s^\star$ for which $\boldsymbol x^\star=\boldsymbol f(\boldsymbol s^\star)$ represents a desired system state, i.e., a physiological gait cycle in the considered application.

\subsubsection*{Control Law and Conceptual Idea} 
The method is based on a reward model, reflecting the control objective, and a steady-state model, associating a feature vector with an input setting.
The reward model is a function that assigns a scalar value to the feature vector estimating an expert rating of the "goodness" of the feature vector.
The reward thereby provides a direction of improvement for the feature vector, which is mapped to a change in input settings via the steady-state model.

We define the control law in terms of input adaptations $\Delta \boldsymbol s$:
\begin{align*}
\Delta \boldsymbol s  = \boldsymbol f^{-1}(\alpha \Delta \boldsymbol x + \boldsymbol x)- \boldsymbol s,
\end{align*}
where $\Delta \boldsymbol x$ is the direction of improvement, $\boldsymbol f^{-1}:\mathbb{R}^n\rightarrow \mathbb{R}^m$ is the steady-state model (the inverse mapping of $\boldsymbol f$ in \eqref{eq:sys}), and $\alpha>0$ is the gain of the control law.
We compute $\Delta \boldsymbol x$ as the gradient of the reward model $r(\boldsymbol x)\in \mathbb{R}$, i.e.,
\begin{align*}
\Delta \boldsymbol x = \nabla_{\boldsymbol x} r(\boldsymbol x).
\end{align*}

\begin{figure}[t]
      \centering
     \includegraphics[width=0.6\columnwidth]{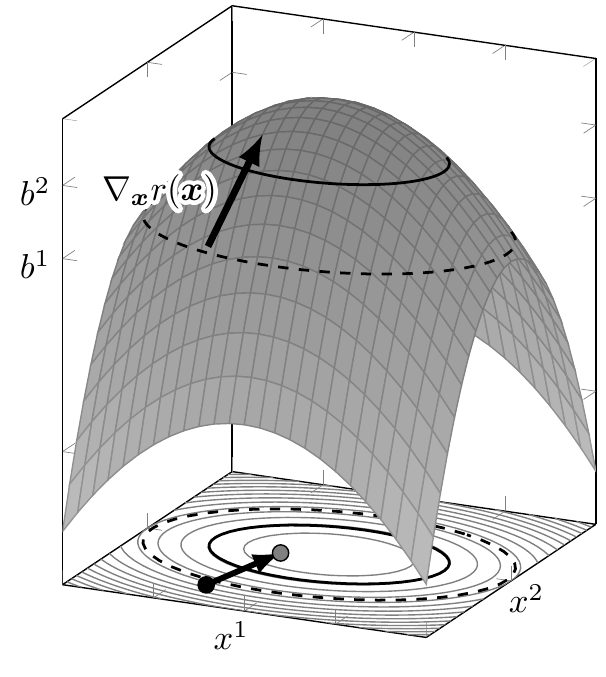}
     \includegraphics[width=0.9\columnwidth]{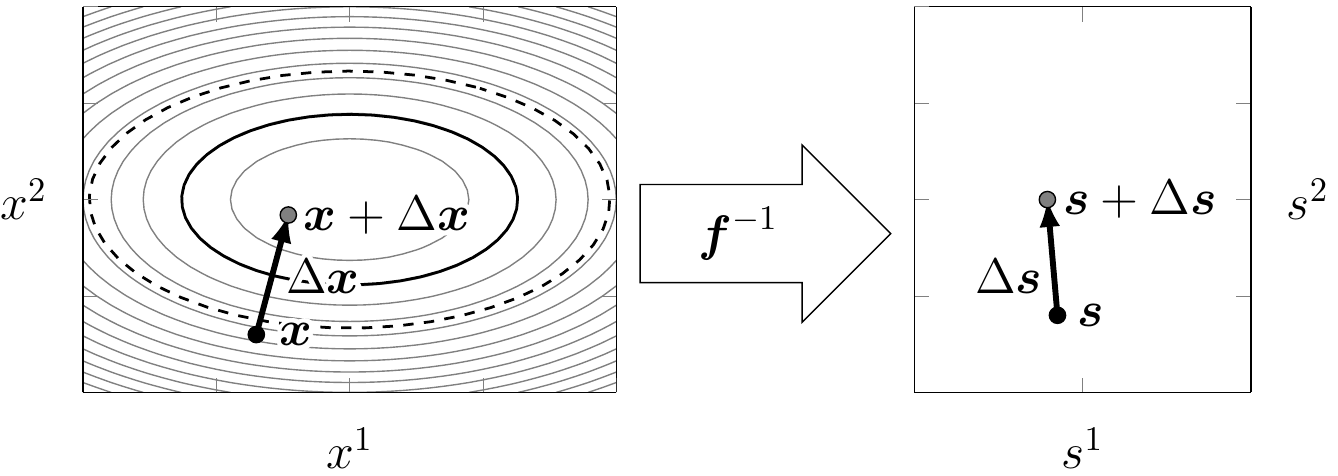}
      \caption{
      Top: Example of reward model with gradient vector $\nabla_{\boldsymbol x} r(\boldsymbol x)$ where $\boldsymbol x=[x^1\ x^2]^\tp$ and projected level sets onto the $x^1-x^2$ plane. 
      The example shows a case of three ratings $r_i=1,2,3$ separated by two classification boundaries indicated as solid black and dashed black ellipses.
  	  Bottom: Steady-state model to compute $\Delta \boldsymbol s$ from $\Delta \boldsymbol x$ where $\boldsymbol s=[s^1\ s^2]^\tp$. 
  }
      \label{fig:reward}
\end{figure}

Fig.~\ref{fig:reward} shows an example of a reward model and indicates how its gradient is used for feedback control using the steady-state model.
Both models $r(\boldsymbol x)$ and $\boldsymbol f^{-1}(\cdot)$ are inferred from data. 
In order to train the reward model, we utilize ratings on an integer scale as samples of the reward model, i.e., $r_i=1,...S$, where $r_i=1$ is the worst and $r_i=S$ is the best rating.
Additionally, we train a steady-state model $\boldsymbol f^{-1}(\cdot)$ to relate the direction of improvement suggested by the reward model to the corresponding input adaptation (bottom part of Fig.~\ref{fig:reward}).
In order to build both the reward model and the steady-state model, $N$ training samples are collected. 
Each training sample with index $i$ consists of a feature vector $\boldsymbol x_i$, the input settings $\boldsymbol s_i$, and the corresponding rating $r_i\in \{1,...,S\}$:
\begin{align}
\label{eq:data}
\{\boldsymbol x_i,\boldsymbol s_i, r_i\}_{i=1}^N.
\end{align}
Note that throughout this paper, the feature vector $\boldsymbol x$ is normalized using collected data $\boldsymbol x_i$ such that the collected data are zero-mean with unit-variance in order to account for different value ranges and units, cf., \cite{Bishop2006}.

\subsubsection*{Outline}
The reward model is trained with the feature vector $\boldsymbol x_i$ and its corresponding rating $r_i$ in \eqref{eq:data} using a supervised learning technique (Section~\ref{sec:model}).
The resulting reward model is then used to compute the gradient $\nabla_{\boldsymbol x} r(\boldsymbol x)$ as direction of improvement. 
Finally, a steady-state model relates this direction of improvement with necessary changes in input settings $\boldsymbol s$. 
The steady-state model is computed using a regression technique (Section~\ref{sec:mapping}).

\subsection{Reward Model using Supervised Learning}
\label{sec:model}
The first step of the framework is the learning of a reward model reflecting the physiology of the gait based on supervised learning techniques \cite{Bishop2006}.
The reward model is a continuous function, i.e., it provides a reward for all $\boldsymbol x$, whereas observations $\boldsymbol x_i$ are potentially sparse. 

In view of the considered application, we postulate a reward model of the form: 
\begin{align}
\label{eq:reward}
r(\boldsymbol x) = 0.5 \boldsymbol x^\tp \boldsymbol W \boldsymbol x + \boldsymbol w^\tp \boldsymbol x + b,
\end{align}
where $\boldsymbol W=\boldsymbol W^\tp\prec 0$, $\boldsymbol w\in \mathbb{R}^n$, and $b\in \mathbb{R}$ are the parameters to be learned from expert ratings given in the form of integers on a scale from $1$ to $S$.
The rationale for selecting a quadratic model with negative definite $\boldsymbol W$ is the observation that the gait degrades in all relative directions when changing input settings from a physiological gait.
Important properties of this reward model are that a vanishing gradient indicates that global optimality has been reached and its computational simplicity.
This motivates the gradient ascent method for optimizing performance. 

In order to learn $\boldsymbol W$, $\boldsymbol w$, and $b$ in \eqref{eq:reward}, we construct $S-1$ classification problems. 
These $S-1$ classification problems share the parameters $\boldsymbol W$, $\boldsymbol w$, and $b$ of the reward model and the corresponding classification boundaries are given by
$$
r^l(\boldsymbol x)=0.5\boldsymbol x^\tp \boldsymbol W\boldsymbol x+\boldsymbol w^\tp \boldsymbol x + b^l
$$ 
for all $l=1,...,\ S-1$ with $b^l=b-l-0.5$ separating the $S$ different ratings such that $r^l(\boldsymbol x_i)>0$ if $r_i>l + 0.5$.
Further, for each data sample $i$ and each $l$, we define 
\begin{align*}
y_i^l = 
\begin{cases}
1 & {\rm if}\ r_i>l + 0.5
\\
-1 & {\rm else}.
\end{cases}
\end{align*}
Hence, an ideal reward model with perfect data and separation satisfies
\begin{align}
\label{eq:perfect}
 y_i^{l} 
 r^l(\boldsymbol x_i) 
 \geq 0
\quad 
\begin{matrix}
\forall\ i=1,...,N 
\\ 
\forall\ l=1,...,S-1.
\end{matrix}
\end{align}
In order to allow for noisy data and imperfect human feedback, \eqref{eq:perfect} is relaxed to find $r^l(\boldsymbol x)$ that satisfies \eqref{eq:perfect} "as closely as possible" by introducing a margin $\xi_i^l\geq 0$. 
This approach is closely related to a Support Vector Machine, cf., \cite{Bishop2006}, with a polynomial kernel function of degree two.
The functions $r^l(\boldsymbol x)$ correspond to $S-1$ classification boundaries in a multi-category classification framework.
The parameters $\boldsymbol W$, $\boldsymbol w$, and $b$ of the reward model \eqref{eq:reward} are computed by solving the following optimization problem using L1 regularization:
\begin{subequations}
\label{eq:classification}
\begin{align}
 \underset{{\boldsymbol W,\boldsymbol w,b^{l},b,\xi_i^l}}{\text{minimize}}\quad  & \sum_{l=1}^{S-1} \sum_{i=1}^{N} \xi_i^ {l} 
+ \lambda_1\cdot 
\left(
\|\boldsymbol W\|_1
+ 
\|\boldsymbol w\|_1
\right)
\\
\text{subject to}\quad &
y_i^{l}
r^l(\boldsymbol x_i)
 \geq 1 -\xi_i^{l},
 \quad  \quad 
\forall\ i=1,...,N 
\\ 
& 
\xi_i^{l}\geq 0, 
\quad \quad  \quad  \quad  \quad 
\forall\ l=1,...,S-1
\\ & 
r^l(\boldsymbol x_i)
=0.5 \boldsymbol x^\tp_i \boldsymbol W\boldsymbol x_i
+\boldsymbol w^\tp \boldsymbol x_i
 +b^{l}
\\ &  b^l=b-l-0.5
\\ & \boldsymbol W=\boldsymbol W^\tp \prec 0
\end{align}
\end{subequations}
where $\lambda_1>0$ controls the trade-off between minimizing the training error and model complexity captured by the norm $\| \boldsymbol W \|_1=\sum_{j=1}^{n} \sum_{k=1}^{n} |W_{jk}|$ (elementwise 1-norm) and $\| \boldsymbol w \|_1$, which is generally applied to avoid overfitting of a model and is sometimes also called lasso regression \cite{Bishop2006}.

\subsection{Feedback Control using Reward Model}
\label{sec:mapping}
The second step of the proposed framework is to exploit the trained reward model for feedback control.
The idea is \textit{(i)} to use the gradient of the reward model as the direction of improvement and \textit{(ii)} to relate this gradient to a desired change in inputs with a steady-state model.
\subsubsection*{(i) Gradient of reward model}
The gradient of the inferred reward model is the direction of best improvement.
The control strategy is to follow this gradient in order to maximize reward. 
The gradient of the quadratic reward model in \eqref{eq:reward} is 
\begin{align*}
\Delta \boldsymbol x = \nabla_{\boldsymbol x} r(\boldsymbol x) = \boldsymbol W \boldsymbol x + \boldsymbol w.
\end{align*}

\subsubsection*{(ii) Mapping of gradient to setting space with steady-state model}
In order to advance the system along the gradient direction, we relate the direction of improvement $\Delta \boldsymbol x$ to a change in input settings with a steady-state model $\boldsymbol f^{-1}$.
We use a first order approximation $\boldsymbol s
= \boldsymbol f^{-1}(\boldsymbol x)
\approx \boldsymbol M\boldsymbol x + \boldsymbol m$ with $\boldsymbol M\in \mathbb{R}^{m\times n}$, $\boldsymbol m\in \mathbb{R}^{m}$ to compute the change in input settings $\Delta \boldsymbol s$ as
\begin{align}
\label{eq:deltaS}
\Delta \boldsymbol s = 
\boldsymbol M( \alpha 
(\boldsymbol W \boldsymbol x  +  \boldsymbol w )+ \boldsymbol x  )
+  \boldsymbol m
-  \boldsymbol s,
\end{align}
where $\alpha$ can be interpreted as feedback gain or the learning rate in a gradient ascent method. 
The steady-state model is estimated as the least squares solution of the data in \eqref{eq:data}:
\begin{align}
\label{eq:mapping}
\underset{\boldsymbol M,\boldsymbol m}{\text{minimize}} 
\sum_{i=1}^{N}\|\boldsymbol s_i - \boldsymbol M\boldsymbol x_i - \boldsymbol m \|_2^2 
+ \lambda_2
\cdot 
(\|\boldsymbol M\|_1 +\|\boldsymbol m\|_1)
\end{align}
where, again, we use $\lambda_2>0$ to control the trade-off between model fit and model complexity.

Using the quadratic reward model in \eqref{eq:reward} and the linear steady-state model in \eqref{eq:mapping}, the application of the proposed control strategy \eqref{eq:deltaS} requires only matrix-vector multiplications, which is computationally inexpensive and can be performed online, cf., Algorithm~\ref{alg} for an overview of the method.
Additionally, as will be shown empirically, the application requires only few online input adaptations.

\begin{algorithm}[!h]
\caption{Training and Application of the Method}
\label{alg}
\begin{algorithmic}[1]
\Require \Comment{rating needed}
\State Collect data set in \eqref{eq:data}. 
\State Compute reward model $\boldsymbol W,\boldsymbol w, b$ with \eqref{eq:classification} and steady-state model $\boldsymbol M,\boldsymbol m$ with \eqref{eq:mapping}.
\Ensure \Comment{no rating needed}
\State \textbf{do}
   \State   \quad   Obtain feature vector $\boldsymbol x$ from measurement. 
        \State \quad Apply  
        adaptation
        $\Delta \boldsymbol s 
        = \boldsymbol M(\alpha (\boldsymbol W \boldsymbol x +  \boldsymbol w )+ \boldsymbol x  )  +\boldsymbol m  -\boldsymbol s
        $.
        \State \quad Wait until steady state is reached.
        \State \textbf{while} stopping criterion not fulfilled 
\Comment{cf. Section~\ref{sec:lokomat_law}}
\end{algorithmic}
\end{algorithm}

\begin{remark}
As we will show in the analysis in Section~\ref{sec:results}, the linear mapping $\boldsymbol s\approx\boldsymbol M\boldsymbol x + \boldsymbol m$ yields sufficient accuracy for the considered application.
For more complex systems, one might consider a different steady-state model, e.g., higher order polynomials or a neural network to approximate $\boldsymbol f^{-1}$.
\end{remark}

\begin{remark}
In principle, reinforcement learning could be applied to directly learn physiological settings.
The proposed two-step and model-based method, in contrast, makes use of the higher dimensionality of the feature vector to characterize the gait cycle.
Its key advantage is that less samples are required online and thus, less steps to find physiological settings, which is essential for the considered application.
\end{remark}

\begin{remark}
The proposed method iteratively approaches the optimal settings $\boldsymbol s^\star$ with the gradient ascent method. 
This is important for the considered application to cautiously adapt the input settings of the robot with a human in the loop.
\end{remark}
\begin{remark}
It is also possible to determine the direction of improvement using second-order derivatives of the reward model, e.g., using a Newton-Raphson method. 
However, as numerical second-order derivatives would be more noisy, we choose first-order derivatives, which are simple and yield a more stable estimate of the best (local) improvement.
\end{remark}


\section{Adaptation of Gait Rehabilitation Robot to Walk Patients Physiologically}
\label{sec:lokomat}
In this section, we show how to apply the method presented in Section~\ref{sec:theory} to automatically adapt, or recommend a suitable adaptation, of the Lokomat's input settings in order to walk patients physiologically.
A core element is the reward model that has been built on therapists' ratings and is used to judge the physiology of the gait.
For simplicity, we adjust settings for the left and right leg symmetrically.
This does not pose a problem for the presented study with nondisabled subjects but might be revisited for impaired subjects in future work.

In this work, we focus on physiological walk and exclude the guidance force and the pelvic input settings as they are mainly used for goal-oriented walk \cite{HocomaNotes15}. 
This exclusion is valid for physiological walk where the guidance force and pelvic settings are kept constant at 100\% and 0cm, respectively.
Hence, there are seven input settings that are considered in the application of the method.

\subsubsection*{Safety}
The proposed method is implemented to augment a previously developed safety controller that ensures safe operation of the Lokomat. 
This safety controller intervenes if the input settings exceed nominal ranges for forces and positions of the robot's joints.
An additional contingency controller stops the robot when the deviation of the measured gait trajectory and the desired gait trajectory becomes too large.
In this way, the overall behavior is guaranteed to have the necessary safety requirements for patient and robot, yet among the safe input settings, the ones that improve the gait cycle are chosen.
The reader is referred to \cite{riener2010locomotor} for a more detailed description of the Lokomat's safety mechanisms.

\subsection{Gait Cycle} 
\begin{figure}[t]
      \centering
     \includegraphics[width=0.99\columnwidth]{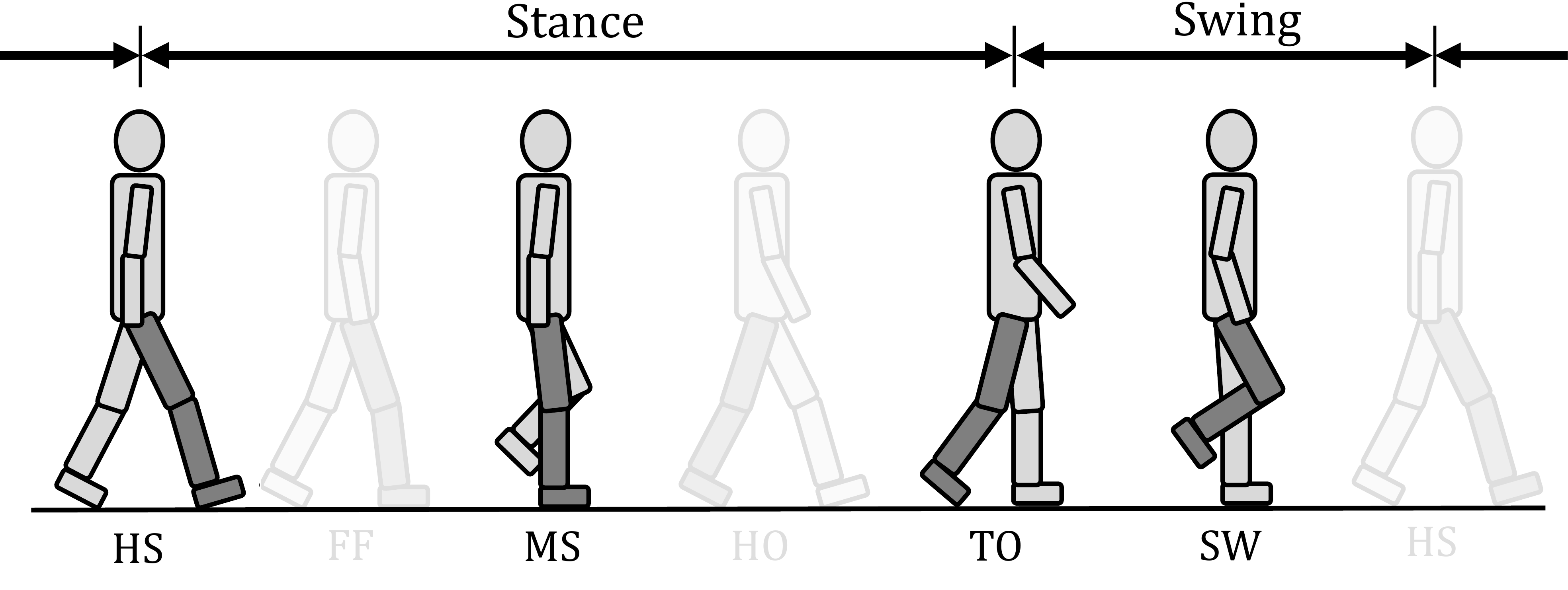}
      \caption{
      Gait phases in order: 
      Heel strike, foot flat (FF), mid-stance, heel off (HO), toe off, mid-swing.
      Both FF and HO phase are not rated in this work, but presented for consistency with the literature \cite{Perry1992}.
      }
      \label{fig:gait}
\end{figure}

The walking of a human is a repetitive sequence of lower limb motions to achieve forward progression.
The gait cycle describes such a sequence for one limb and commonly defines the interval between two consecutive events that describe the heel strike (initial ground contact) \cite{Perry1992}. 
The gait cycle is commonly divided into two main phases, the stance and the swing phase.
The stance phase refers to the period of ground contact, while the swing phase describes limb advancement. 
Fig.~\ref{fig:gait} illustrates the subdivision of these two main phases of the gait cycle into multiple sub-phases, beginning and ending with the heel strike.
This results in a common description of gait using a series of discrete events and corresponding gait phases \cite{Perry1992}.
We focus on four particular phases of the gait cycle, which are emphasized in Fig.~\ref{fig:gait}: 
\begin{itemize}
\item Heel strike (HS): The moment of initial contact of the heel with the ground.
\item Mid-stance (MS): The phase in which the grounded leg supports the full body weight.
\item Toe off (TO): The phase in which the toe lifts off the ground.
\item Mid-swing (SW): The phase in which the raised leg passes the grounded leg.
\end{itemize}

\subsection{Evaluation of Gait Cycle and Data Collection}
We derive the reward model based on the four phases.
For evaluating the four gait phases, we introduce a scoring criterion in consultation with experienced therapists:
\begin{enumerate}[{Rating} 1:]
\item Safe, but not physiological.
\item Safe, not entirely physiological gait cycle.
\item Safe and physiological gait cycle.
\end{enumerate} 
   
\subsubsection*{Data Collection}
A total of 16 nondisabled subjects participated in the data collection.
The 16 subjects were between 158cm - 193cm (5'2'' - 6'4'') in height, 52kg - 93kg (115lbs - 205lbs) in weight, and aged 25 - 62.
Informed consent for the use of the data has been received from all human subjects.
The data collection for each subject involved an evaluation of the four gait phases by therapists for several input settings in order to collect data in a wide range of gait cycles.
The nondisabled subjects were instructed to be passive throughout the data collection, i.e., they were walked by the robot.
This allowed us to collect data for both physiological and non-physiological gait cycles.
Measurements of the Lokomat were recorded for all evaluations.
For each subject, the experienced therapists first manually tuned the input settings to achieve rating 3 for all four phases (Set~0 in Table~\ref{tb:experimentalsetup}), where the resulting input settings are referred to as initial physiological gait (IPG).
Table~\ref{tb:experimentalsetup} shows the input settings used in the data collection as deviations from the initial physiological gait.
Each subject walked for approximately 60 seconds for each set of input settings, while the therapist provided evaluations of the walking pattern. 
The assessment started after a transient interval of approximately 15 seconds to ensure that the walking has reached a steady state.
Note that the input settings resulting in a physiological gait pattern varied between subjects. 
\begin{table}[t]
\setlength{\tabcolsep}{3pt}
\caption{
Input Setting for Data Collection
}
\begin{center}
\def\arraystretch{1.00}
\label{tb:experimentalsetup}
\begin{tabular}{rll } 
\hline
Set & Input Settings & Value\\
\hline
\vspace{-0.25cm}
\\
\arabic{SetupCount}
\stepcounter{SetupCount} 
& Initial Set & IPG
\\
		\arabic{SetupCount}
\stepcounter{SetupCount} 
& Hip Range of Motion & IPG  + 6${}^\circ$\\
		\arabic{SetupCount}
\stepcounter{SetupCount} 
& Hip Range of Motion & IPG  + 12${}^\circ$\\	
		\arabic{SetupCount}
\stepcounter{SetupCount} 
& Hip Range of Motion & IPG  -- 6${}^\circ$\\
		\arabic{SetupCount}
\stepcounter{SetupCount} 
& Hip Range of Motion & IPG  -- 12${}^\circ$\\	
		\arabic{SetupCount}
\stepcounter{SetupCount} 
& Hip Offset & IPG  + 4${}^\circ$\\
		\arabic{SetupCount}
\stepcounter{SetupCount} 
& Hip Offset & IPG  + 8${}^\circ$\\
		\arabic{SetupCount}
\stepcounter{SetupCount} 
& Hip Offset & IPG  -- 5${}^\circ$\\
		\arabic{SetupCount}
\stepcounter{SetupCount} 
& Hip Range of Motion, Hip Offset & IPG  + 12${}^\circ$ /  IPG  -- 3${}^\circ$\\
		\arabic{SetupCount}
\stepcounter{SetupCount} 
& Hip Range of Motion, Hip Offset & IPG  + 12${}^\circ$ /  IPG  + 3${}^\circ$\\
		\arabic{SetupCount}
\stepcounter{SetupCount} 
& Hip Range of Motion, Hip Offset & IPG  -- 12${}^\circ$ /  IPG  -- 3${}^\circ$\\
		\arabic{SetupCount}
\stepcounter{SetupCount} 
 & Hip Range of Motion, Hip Offset & IPG  -- 12${}^\circ$ /  IPG  + 3${}^\circ$ \\
		\arabic{SetupCount}
\stepcounter{SetupCount} 
& Knee Range of Motion & IPG  + 6${}^\circ$\\
		\arabic{SetupCount}
\stepcounter{SetupCount} 
& Knee Range of Motion & IPG  + 12${}^\circ$\\
		\arabic{SetupCount}
\stepcounter{SetupCount} 
& Knee Range of Motion & IPG  -- 9${}^\circ$\\
		\arabic{SetupCount}
\stepcounter{SetupCount} 
& Knee Range of Motion & IPG  -- 15${}^\circ$\\
		\arabic{SetupCount}
\stepcounter{SetupCount} 
& Knee Offset & IPG  + 4${}^\circ$ \\
		\arabic{SetupCount}
\stepcounter{SetupCount} 
& Knee Offset & IPG  + 8${}^\circ$ \\
		\arabic{SetupCount}
\stepcounter{SetupCount} 
& Knee Range of Motion / Knee Offset & IPG  + 15${}^\circ$ / IPG  + 6${}^\circ$\\
		\arabic{SetupCount}
\stepcounter{SetupCount} 
& Knee Range of Motion / Knee Offset & IPG  + 21${}^\circ$ /  IPG  + 6${}^\circ$ \\
\arabic{SetupCount}
\stepcounter{SetupCount} 
& Speed & IPG + 0.5km/h\\
		\arabic{SetupCount}
\stepcounter{SetupCount} 
& Speed & IPG + 1.0km/h\\
		\arabic{SetupCount}
\stepcounter{SetupCount} 
& Speed & IPG -- 0.5km/h\\
		\arabic{SetupCount}
\stepcounter{SetupCount} 
& Speed & IPG -- 1.0km/h\\
\arabic{SetupCount}
\stepcounter{SetupCount} 
& Orthosis Speed &   IPG  + 0.03 \\
		\arabic{SetupCount}
\stepcounter{SetupCount} 
& Orthosis Speed & IPG  + 0.05 \\
		\arabic{SetupCount}
\stepcounter{SetupCount} 
& Orthosis Speed & IPG  -- 0.03 \\
		\arabic{SetupCount}
\stepcounter{SetupCount} 
& Orthosis Speed & IPG  -- 0.05 \\
		\arabic{SetupCount}
\stepcounter{SetupCount} 
& Bodyweight Support & IPG  + 15\%\\
		\arabic{SetupCount}
\stepcounter{SetupCount} 
& Bodyweight Support & IPG  + 30\%\\
		\arabic{SetupCount}
\stepcounter{SetupCount} 
& Bodyweight Support & IPG  -- 15\%\\
		\arabic{SetupCount}
\stepcounter{SetupCount} 
& Bodyweight Support & IPG  -- 30\%\\
\hline
\end{tabular}
\end{center}
\end{table}

The scoring criterion and the consideration of the four phases, as well as the data collection protocol were introduced in consultation with clinical experts from Hocoma (U. Costa and P. A. Gon\c calves Rodrigues, personal communication, Nov. 05, 2017).
As a result, we obtained the chosen input settings, the corresponding ratings on an integer scale from 1 to 3, and the recording of measurements of the Lokomat.
Next, we discuss the computation of the feature vector from the recorded measurements.

\subsubsection*{Feature Vector}
\begin{figure}[t]
      \centering
     \includegraphics[width=0.9\columnwidth]{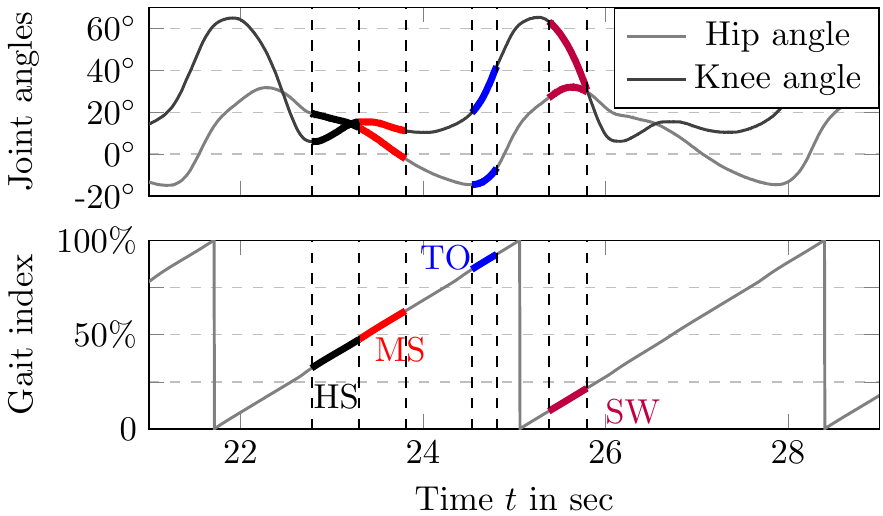}
      \caption{Top: Joint angles. Bottom: Segmentation of time signals into four phases using the gait index with HS in 34.5\%-47.5\%, MS in 47.5\%-65.5\%, TO in 84.5\%-92.5\%, and SW in 9.5\%-21.5\% of one period of the gait index. 
The falling edge of the gait index does not align with the biomechanical definition of a gait cycle but enables separation of the gait cycle into phases.
}
      \label{fig:gaitindex}
\end{figure}

We use the gait index signal of the Lokomat as an indicator to identify progression through the gait cycle.
The gait index is a sawtooth signal and is displayed in the bottom plot in Fig.~\ref{fig:gaitindex}.
It is used to determine the time-windows of the four phases, cf., the dashed lines in Fig.~\ref{fig:gaitindex}.
The time-windows are used to compute the feature vector, composed of statistical features for power, angle, and torque for both hip and knee joints, cf., Table~\ref{tb:feature}.
The result is one feature vector for each phase: $\boldsymbol x_{\rm HS}, \boldsymbol x_{\rm MS}, \boldsymbol x_{\rm TO}, \boldsymbol x_{\rm SW} \in \mathbb{R}^{12}$.
The Lokomat provides measurements of all the signals listed in Table~\ref{tb:feature} synchronized by the gait index signal, which makes the computation of the features simple.
\begin{table}[t]
\caption{
Values for Feature Vector
}
\begin{center}
\def\arraystretch{1.00}
\label{tb:feature}
\begin{tabular}{lccccc }
\hline
\# & Joint & Signal & Unit &  Feature      \\
\hline
\\
\vspace{-0.575cm}
\\
$x^1$ & hip &
joint power
& Nm/s
& mean 
\\
$x^2$ & hip &
angle
& rad
& min 
\\
$x^3$ & hip &
angle
& rad
& max 
\\
$x^4$ & hip &
angle
& rad
& range 
\\
$x^5$ & hip &
torque
& Nm
& mean 
\\
$x^6$ & hip &
torque
& Nm
& variance 
 \\
$x^7$ & knee &
joint power
& Nm/s
& mean 
\\
$x^8$ & knee &
angle
& rad
& min 
\\
$x^9$ & knee &
angle
& rad
& max 
\\
$x^{10}$ & knee &
angle
& rad
& range 
\\
$x^{11}$ & knee &
torque
& Nm
& mean 
\\
$x^{12}$ & knee &
torque
& Nm
& variance 
\\
\hline
\end{tabular}
\end{center}
\end{table}

\begin{remark}
For each subject, the data collection takes around one hour, including rating the gait cycle.
As described in Algorithm~\ref{alg}, the application of the control law does not include further training and the control law is therefore not personalized to the subject.
\end{remark}
\begin{remark}
Initially, we defined more features than the twelve in Table~\ref{tb:feature}, e.g., frequency domain features, which the supervised learning problem in \eqref{eq:classification} with L1 regularization discarded.
In order to reduce the problem dimension in the online algorithm, we discarded them as well.
\end{remark}

\subsection{Reward Model and Steady-State Model for Lokomat}
Given the data set, we apply the method in Section~\ref{sec:theory} to learn four reward models.
We obtain a reward model for each of the four phases represented as $\boldsymbol W_j$, $\boldsymbol w_j$, and $b_j$ from solving \eqref{eq:classification}, where $j\in \{{\rm HS, MS,TO,SW}\}$.

The steady-state model $\boldsymbol M\in \mathbb{R}^{7 \times 48}$, $\boldsymbol m\in \mathbb{R}^{7}$ in \eqref{eq:mapping} is computed by stacking the features of the four phases:
\begin{align*}
\boldsymbol   x = \begin{bmatrix}
    \boldsymbol  x_{\rm HS}^\tp
  &
  \boldsymbol    x_{\rm MS}^\tp
  &
 \boldsymbol     x_{\rm TO}^\tp
  &
 \boldsymbol     x_{\rm SW}^\tp
  \end{bmatrix}^\tp.
\end{align*}

\subsection{Control Law for Gait Rehabilitation Robot}
\label{sec:lokomat_law}
Once the four reward models and the steady-state model are trained using the data in \eqref{eq:data}, the controller automatically chooses input setting adaptations given the current measurements, i.e., it does not require ratings from therapists.
The input adaptation $\Delta \boldsymbol s$ is computed as
\begin{align*}
\Delta \boldsymbol s = 
\boldsymbol M 
\begin{bmatrix}
\alpha  ( \boldsymbol W_{\rm HS}\boldsymbol x_{\rm HS} + \boldsymbol  w_{\rm HS})   + \boldsymbol x_{\rm HS}
  \\
\alpha   (\boldsymbol W_{\rm MS} \boldsymbol x_{\rm MS} + \boldsymbol w_{\rm MS}) +   \boldsymbol x_{\rm MS}
  \\
\alpha  ( \boldsymbol W_{\rm TO}\boldsymbol  x_{\rm TO} +\boldsymbol  w_{\rm TO})   +\boldsymbol  x_{\rm TO}
  \\
\alpha (  \boldsymbol W_{\rm SW}\boldsymbol   x_{\rm SW} +\boldsymbol  w_{\rm SW}) + \boldsymbol   x_{\rm SW}
  \end{bmatrix}
 + \boldsymbol m
  - \boldsymbol s.
\end{align*}

While $\Delta \boldsymbol s$ yields continuous values, the input settings are adjusted in discrete steps, cf., the step-sizes in Table~\ref{tb:settings}.
We aim to change one setting at a time, which is common practice for therapists \cite{HocomaNotes15} and eases the evaluation. 
The following suggests a method to select one single adaptation from $\Delta \boldsymbol s$. 

\subsubsection*{Input Setting Selection \& Stopping Criterion}
In order to select one single discrete change in input setting, we normalize $\Delta \boldsymbol s$ to account for different value ranges and different units per individual setting and select the input corresponding to the largest in absolute value:
\begin{align*}
k^\star =
\arg \max_{k=1,...,7} 
\left| 
\frac{\Delta s^{k}}{\bar s^k - \underbar s^k }
\right|
\end{align*}
with associated index $k^\star$, where the normalization $\bar s^k-\underbar s^k$ is the range of the input setting $k$ in Table~\ref{tb:settings}.
Hence, the algorithm chooses one adaptation with step-size in Table~\ref{tb:settings}.
The input adaptation is stopped when the largest normalized absolute value of change is smaller than a pre-defined parameter $\beta$, i.e.,
$
\left| 
\Delta s^{k^\star}
/ (\bar s^{k^\star} - \underbar s^{k^\star})
\right|
\leq \beta$. 
This indicates closeness to the optimum, i.e., that a physiological gait is reached.
\section{Model Evaluation in Simulation} 
\label{sec:results} 
We first analyze the algorithm in simulation to investigate the model quality.
In this simulation study, we compare two reward models: One that uses ratings on an integer scale from 1 to 3 ($S=3$ in \eqref{eq:classification}) and one that uses only binary ratings, i.e., good and bad ($S=2$  in \eqref{eq:classification}).
For the case $S=3$, we use the collected ratings without modification. 
For the case $S=2$, we combine the data points with rating~1 and rating~2 as \textit{samples of a bad gait} and use the data points with rating~3 as \textit{samples of a good gait}.

\subsection{Evaluation Metrics and Results}
In order to evaluate the trained models, we split the experimentally collected data into training (80\%) and validation data (20\%).
This split is done randomly and repeated 500 times to assess the robustness of the models.
This technique is known as 5-fold cross validation \cite{James2013} and ensures that the validation data are not biased by training on the same data.

\subsubsection{Evaluation of Reward Model}
\label{sec:evalReward}
We evaluate the accuracy of the reward model trained with three ratings, i.e., $S=3$, by computing the pairwise difference in estimated rewards $r(\boldsymbol x_i)-r(\boldsymbol x_j)$ for two data samples $i$ and $j$, classified with respect to their ratings $r_i$ and $r_j$.
The metric is motivated by the fact that two different ratings should be distinguishable.
We define $\Delta \bar r_{nm}$ as
\begin{align}
\label{eq:metric_reward}
\Delta \bar r_{nm}
&=
\frac{1}{|I_n| |I_m|}
\sum_{i\in I_n}\
\sum_{j\in I_m}
\left(
 r(\boldsymbol x_i) -  r(\boldsymbol x_j)\right),
\end{align}
where $I_n=\{i | r_i=n\}$ is an index set of data points with ratings $r_i=n$. 
If the trained reward model and data were perfect, $\Delta \bar r_{nm}=n-m$ with zero standard deviation.

We evaluate the accuracy of the reward model trained with binary ratings, i.e., $S=2$, by computing the classification accuracy of good versus bad ratings:
\begin{align}
\label{eq:metric_reward_binary}
\bar p_{\rm good/bad}
&=
\frac{
1
}{|I_{\rm good}| |I_{\rm bad}|  }
\sum_{i\in I_{\rm good}}\
\sum_{j\in  I_{\rm bad}}
\mathcal{I}_{r(\boldsymbol x_i) >  r(\boldsymbol x_j)}
\end{align}
with $\mathcal{I}_{r(\boldsymbol x_i) >  r(\boldsymbol x_j)}=1$ if $r(\boldsymbol x_i) >  r(\boldsymbol x_j)$ and $\mathcal{I}_{r(\boldsymbol x_i) >  r(\boldsymbol x_j)}=0$, otherwise, and $I_{\rm good}= I_3$ and $I_{\rm bad}=I_{1}\cup I_2=\{i| r_i=1\ {\rm or}\ r_i=2\}$.

\begin{figure}[t]
      \centering
     \includegraphics[width=0.99\columnwidth]{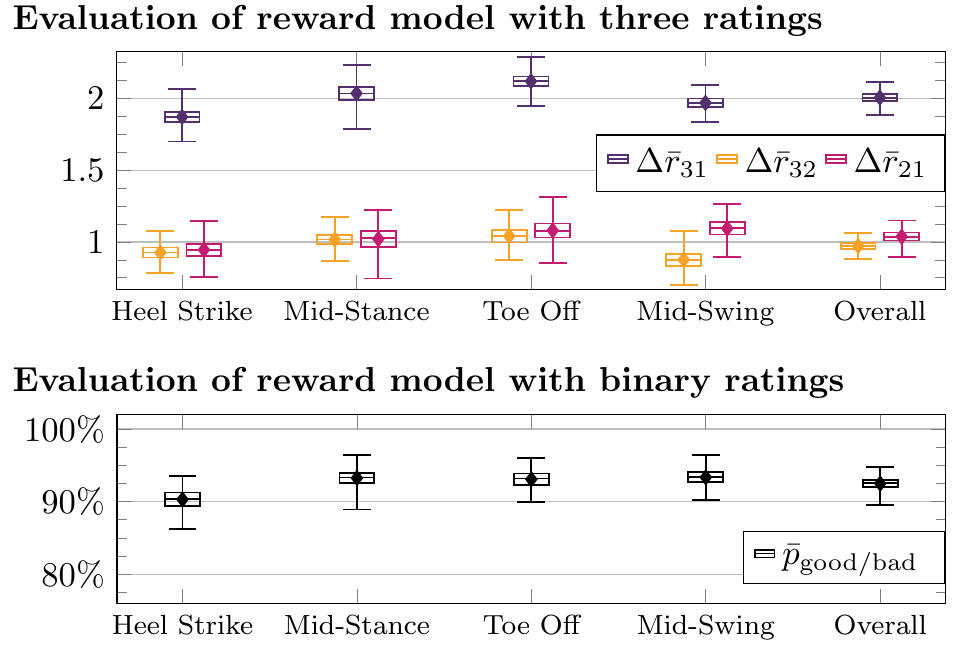}
      \caption{Evaluation of reward models for individual phases and overall.
      The mean over 500 splits of training and validation data, along with the median,  25th  and  75th  percentiles,  and maximum  and  minimum values  are  indicated  by  a diamond, a line,  box  edges,  and whiskers, respectively.
      Top: Pairwise difference in estimated rewards $\Delta \bar r_{31}$, $\Delta \bar r_{32}$, and $\Delta \bar r_{21}$.
      Bottom: Classification accuracy $\bar p_{\rm good/bad}$.
      }
      \label{fig:evalRewardSim}
\end{figure}

Fig.~\ref{fig:evalRewardSim} reports statistical values of both $\Delta \bar r_{nm}$ in \eqref{eq:metric_reward} for evaluating the reward model with $S=3$ and $\bar p_{\rm good/bad}$ in \eqref{eq:metric_reward_binary} for evaluating the reward model with $S=2$ over the 500 splits of training and validation data.
For the reward model computed with $S=3$, the overall deltas in estimated rewards match the deltas in ratings very closely with $2.00$ for $\Delta \bar r_{31}$, $0.97$ for $\Delta \bar r_{32}$, and $1.04$ for $\Delta \bar r_{21}$.
For the reward model computed with $S=2$, the overall classification accuracy ($r(\boldsymbol x_i)>r(\boldsymbol x_j)$ if $r_i>r_j$) is 92.5\%.

\subsubsection{Evaluation of Steady-State Model}
The steady-state model is evaluated using the prediction error $\bar{e}^k$ defined as
\begin{align}
\label{eq:metric_mapping}
\bar{e}^k=\frac{1}{N}\sum_{i=1}^{N} |s_i^k - \boldsymbol M_{k\star} \boldsymbol x_i - \boldsymbol m_{k}|,
\end{align}
where $k$ is the index of the input setting and $\boldsymbol M_{k\star}$ is the $k$th row of matrix $\boldsymbol M$ and $\boldsymbol m_{k}$ is the $k$th entry of vector $\boldsymbol m$.
As we use normalized values for the input settings with $s_i^k \in [0,1]$, the error $\bar{e}^k$ can be interpreted as a percentage offset from the correct input setting.

Table~\ref{tb:evalMapping} reports mean and standard deviation of the errors $\bar e^k$ in \eqref{eq:metric_mapping} over the 500 random splits of training and validation data for all input settings $k$.
It shows an overall average error of 4.17\% and that the errors for all input settings are consistently lower than 6\%.

\subsubsection{Evaluation of Overall Algorithm}
\label{sec:evalOverall}

We evaluate the performance of the overall algorithm by comparing the collected data with the output of the algorithm.
Let the changes in input settings during data collection for all data samples $i=1,...N$ be $\Delta \boldsymbol s^{\rm ex}_i= \boldsymbol s^{\rm ex} - \boldsymbol s_{i}$, where $\boldsymbol s_i$ are the input settings of data point $i$ and $\boldsymbol s^{\rm ex}$ are the physiological settings, which are set by the therapist at the beginning of the data collection.
Note that $\boldsymbol s^{\rm ex}$ depends on the subject, however, we omit this dependency in the notation for ease of exposition.
It is also important to note that $\boldsymbol s^{\rm ex}$ are not the only possible physiological input settings.
We compare the input adaptation proposed by our algorithm $\Delta \boldsymbol s_i$ against the deviation from the physiological settings $\Delta \boldsymbol s_i^{\rm ex}$, where we can have three different outcomes: 
\subsubsection*{Case 1 (Same Setting \& Same Direction)}
The algorithm selects the input adaptation in the same direction as during data collection, 
which is known to be a correct choice as it is closer to the physiological settings $\boldsymbol s^{\rm ex}$.
\subsubsection*{Case 2 (Same Setting \& Opposite Direction)}
The algorithm selects the same setting but in the opposite direction as during data collection, 
which is likely to be an incorrect choice.
\subsubsection*{Case 3 (Different Setting)}
The algorithm selects a different input adaptation, the implications of which are unknown and could be either correct or incorrect, which cannot be evaluated without closed-loop testing.

\begin{table}[t]
\caption{
Mean and Standard Deviation of Steady-State Model
}
\begin{center}
\def\arraystretch{1.02}
\label{tb:evalMapping}
\begin{tabular}{llllll }
\hline
$s^k$ & Setting & Error  $\bar{e}^k$ \\
\hline
\\
\vspace{-.6cm}
\\
$s^1$ & Hip Range of Motion  & $0.0578 \pm 0.0019$  
\\
$s^2$  & Hip Offset & $0.0370 \pm 0.0008$ \\
$s^3$  &  Knee Range of Motion & $0.0547 \pm 0.0021$ \\
$s^4$  &  Knee Offset  & $0.0324 \pm 0.0009$\\
$s^5$ & Speed & $0.0307 \pm 0.0009$ \\
$s^6$  &   Orthosis Speed  & $0.0315 \pm 0.0010$ \\
$s^7$ &  Bodyweight Support& $0.0542 \pm 0.0017$ \\
\hline 
 & Overall & $0.0417 \pm 0.0186$
 \\
\hline
\end{tabular}
\end{center}
\end{table}

We compute the percentage of data points falling in each case for each setting $k$ and for $\Delta \boldsymbol s_i^{\rm ex} = 0$ (no adaptation), i.e., $p_{\rm C1}^k$, $p_{\rm C2}^k$, and $p_{\rm C3}^k$ for Case~1, Case~2, and Case~3, respectively, where $p_{\rm C1}^k+p_{\rm C2}^k+p_{\rm C3}^k=1$.
If the algorithm replicated the data collection perfectly, then $p_{\rm C1}^k=1$ for all settings $k$.
Given the discrete and unique setting selection, the overall algorithm has 15 options to choose from: An increase in one of the seven settings by one unit, a decrease in one of the seven settings by one unit, or \textit{no adaptation}.
Hence, random decision-making yields a probability of $p=1/15\approx 6.7\%$ for each option.

Table~\ref{tb:simulation} reports mean and standard deviation of the percentage values of the three cases.
The algorithm chooses the input adaptations for hip range of motion, hip offset, knee range of motion, and knee offset very often when their adaptation leads to $\boldsymbol s^{\rm ex}$ ($86.7\%$--$100.0\%$).
Also, it often chooses \textit{no adaptation} when the gait is physiological, with input settings $\boldsymbol s^{\rm ex}$.
Table~\ref{tb:simulation} also shows that decision-making with the proposed algorithm is more ambiguous for the input adaptations of speed, orthosis speed, and bodyweight support.
Overall, the algorithm proposes a setting that is closer to $\boldsymbol s^{\rm ex}$ (Case 1) in $80.7\%$ and $80.6\%$ for the reward models trained with $S=3$ and $S=2$, respectively.
The algorithm suggests a probably incorrect input adaptation in less than $1\%$  (Case 2).  
In around $19\%$, the algorithm suggests a different input adaptation (Case 3).

\begin{table}[t]
\setlength{\tabcolsep}{4pt}
\caption{
Evaluation of Overall Algorithm in Simulation
}
\begin{center}
\def\arraystretch{1.0}
\label{tb:simulation}
\begin{tabular}{ll rrrr} 
\hline
 \multicolumn{5}{c}{\textbf{Three ratings (1, 2, or 3) $S=3$}} 
\\
  $s^k$ 
 & 
 Setting  & 
 $p_{\rm C1}^k$ in \%   & 
 $p_{\rm C2}^k$ in \%    & 
 $p_{\rm C3}^k$ in \%   
 \\
 \vspace{-0.25cm}
 \\
\hline
\vspace{-0.25cm}
\\
& No Adaptation 
 &  $77.6\pm 3.6$ & -  & $22.4\pm 3.6$
 \\ 
 $s^1$ & Hip Range of Motion  & $86.7 \pm 1.8$   & $0$  & $13.3\pm 1.8$
 \\
 $s^2$ & Hip Offset  & $96.4\pm 1.0$ & $0$ & $3.6\pm 1.0$
 \\ 
 $s^3$  & Knee Range of Motion  & $91.0 \pm 1.7$ & $0$ & $9.1\pm 1.7$
 \\
 $s^4$ & Knee Offset  & $100.0\pm 0.0$ & $0$  & $0$
 \\
 $s^5$ & Speed  & $71.1 \pm 2.8$ & $0$  & $29.0\pm 2.8$
 \\
 $s^6$ & Orthosis Speed  & $33.3 \pm 3.7$ & $3.5 \pm 1.6$ & $63.2\pm 3.8$
 \\ 
 $s^7$ & Bodyweight Support  & $55.6\pm 3.8$ & $0$  & $44.5\pm 3.8$
 \\
\hline 
& Overall accuracy & $80.7 \pm 1.0$ &  $0.3 \pm 0.1$   &  $19.0\pm 1.0$
 \\
\hline
\\
\vspace{-.5cm}
\\
\multicolumn{5}{c}{\textbf{Binary ratings (good or bad) $S=2$}}
\\
  $s^k$ 
 & 
 Setting  & 
 $p_{\rm C1}^k$ in \%   & 
 $p_{\rm C2}^k$ in \%   & 
 $p_{\rm C3}^k$ in \%   
 \\
 \vspace{-0.25cm}
 \\
\hline
\vspace{-0.25cm}
\\
& No Adaptation 
 &  $76.8\pm 3.7$ & -  &  $23.2\pm 3.7$
 \\ 
 $s^1$ & Hip Range of Motion  
 & $88.6 \pm 1.7$   & $0$  &  $11.4\pm 1.7$
 \\
 $s^2$ & Hip Offset  
 & $95.6\pm 1.1$ & $0$  &  $4.4\pm 1.1$
 \\ 
 $s^3$  & Knee Range of Motion  
 & $90.0 \pm 2.0$ & $0$ &  $10.0\pm 2.0$
 \\
 $s^4$ & Knee Offset  
 & $100.0\pm 0.0$ & $0$  & $0$
 \\
 $s^5$ & Speed  
 & $71.3 \pm 2.9$ & $0$  &  $28.7\pm 2.9$
 \\
 $s^6$ & Orthosis Speed  
 & $32.1 \pm 3.8$ & $2.9 \pm 1.4$  &  $65.0\pm 3.9$
 \\ 
 $s^7$ & Bodyweight Support  
 & $53.9\pm 3.9$ & $0$  & $46.2\pm 3.9$
 \\
\hline 
& Overall accuracy 
& $80.6 \pm 1.1$ &  $0.2 \pm 0.1$  &  $19.2\pm 1.0$
 \\
\hline
\end{tabular}
\end{center}
\end{table}

\begin{remark}
The same evaluation using exclusively kinematic features ($x^2,x^3,x^4,x^8,x^9,x^{10}$ in Table~\ref{tb:feature}) yields slightly different results with overall $p_{C1}=79.1\%\pm 1.0\%$, $p_{C2}=0.2\%\pm 0.1\%$, and $p_{C3}=20.6\%\pm 1.0\%$ for binary ratings.
A purely kinematic feature vector might be important when working with impaired patients, where power/torque features might be an indication of individual impairments rather than a characterization of a physiological gait.
\end{remark}

\subsection{Discussion}
The rewards predicted with the reward model trained with three ratings (two classification boundaries at $1.5$ and $2.5$), match the true ratings very closely. 
The reward model trained with binary ratings (one classification boundary at $2.5$) is able to distinguish \textit{good} from \textit{bad} gait patterns confidently with an overall classification accuracy of more than $90\%$.
The steady-state model shows an average error of 5\%.
As we will show in Section~\ref{sec:results_ex}, this accuracy suffices for the considered application.
For example, the expected error of 3.07\% of $s^5$ translates into an error in treadmill speed of 0.075m/s and the expected error of 5.78\% of $s^1$ translates into an error in hip range of motion of 2.08${}^\circ$, which is less than one input setting step-size, cf., Table~\ref{tb:settings}.
Even though another model may increase accuracy, it may come at the expense of increased complexity in the computation. 
Our linear model only requires matrix-vector multiplication, which can easily be implemented on the controller of the Lokomat and is chosen as a suitable compromise of simplicity and accuracy.

The evaluation of both components, the reward model and the steady-state model, in simulation allow us to conclude that they provide suitable models for the considered application.
For the overall algorithm, Case 1 is known to result in an improved physiology of the gait cycle. 
Case 3, however, does not imply that the suggested adaptation will not lead to an improved gait cycle as there may be multiple different input adaptations that lead to a physiological gait (not only $\boldsymbol s^{\rm ex}$).
In these cases, we do not know if the suggested adaptation would have led to an improvement in gait without closed-loop testing.
Hence, the probabilities  $80.7\%$ and $80.6\%$ of Case~1 for the two reward models can be interpreted as a lower bound for the overall improvement.
The relatively low standard deviation for all settings indicates that the learning is robust against variation in the training data.
The use of binary ratings eases the data collection and has been shown to perform similarly well. 
Therefore, we proceed with closed-loop testing of the algorithm using a reward model trained with binary ratings.

\section{Experimental Results - Closed-Loop Testing}
\label{sec:results_ex}
The proposed algorithm was implemented as a recommendation system on the Lokomat for closed-loop evaluation.
We implemented the algorithm using the reward model trained with binary ratings (good and bad) of the gait cycle.
It is important to note, that no data from the respective test person was used for training of the reward or the steady-state model.

\subsection{Experiment Setup}
We conducted experiments with ten nondisabled subjects and ten different sets of initial non-physiological gait cycles (test scenarios).
Table~\ref{tb:experiment} describes the ten test scenarios and outlines input adaptations that therapists are expected to make (according to the heuristic guidelines).
Scenario~1 through 8 are very common observations of a patient's gait cycle on the Lokomat.
Scenario~9 and 10 are combinations of the more common flaws and are included to challenge the algorithm with more complex scenarios.
\begin{table}[t]
\setlength{\tabcolsep}{2pt}
\caption{
Test Scenarios of Experiment
}
\begin{center}
\def\arraystretch{1.0}
\label{tb:experiment}
\begin{tabular}{rllll}
\hline	 
\multicolumn{2}{l}{Scenario:  Observations}  
&
Therapists' heuristic rules (expectation) 
\\
\hline
\vspace{-0.25cm}
\\
1: & 
Limited foot clearance, & 
Increase knee range of motion 
\\
& foot dropping & ($s^3 \uparrow$)

 \\
2: & Short steps & Increase hip range of motion, speed 
\\
  &   & ($s^1 \uparrow$, $s^5 \uparrow$)
\\
3: & Foot dragging & Decrease speed, increase   orthosis speed
\\
 &  &  ($s^5 \downarrow$, $s^6 \uparrow$)
 \\
4: & Large steps, & Decrease hip range of motion, hip offset
\\
 &  late heel strike &  ($s^1 \downarrow$, $s^2 \downarrow$)
 \\
5: & Short steps,   & Increase hip range of motion,  hip offset
 \\
 & hip extension &  ($s^1 \uparrow$, $s^2 \uparrow$)
 \\
6: & Bouncing & Decrease speed, bodyweight support
 \\
 &   &  ($s^5 \downarrow$, $s^7 \downarrow$)
 \\
7: & Foot slipping & Decrease knee range of motion,  
 \\
  &   &  orthosis speed ($s^3 \downarrow$, $s^6 \downarrow$)
 \\
8: & Knee buckling & Increase knee range of motion,  
 \\
 &   &  bodyweight support ($s^3 \uparrow$ ,$s^7 \uparrow$)
 \\
9: & Large steps,   & Decrease hip range of motion, increase hip  
 \\
 &  early heel strike &  offset, increase speed ($s^1 \downarrow$, $s^2 \uparrow$, $s^5 \uparrow$)
 \\
10: & Large steps, late heel   & Decrease hip range of motion, hip offset, 
 \\
 &   strike, foot slipping &  knee range of motion, orthosis speed  
 \\
 &   &   ($s^1 \downarrow$, $s^2 \downarrow$, $s^3 \downarrow$, $s^6 \downarrow$)
 \\
\hline
\end{tabular}
\end{center}
\end{table}

The selected scenarios cover the most common observations of the gait cycle of a passive subject (without muscle activity) on the Lokomat and, therefore, are expected to adequately evaluate the proposed algorithm experimentally (with nondisabled subjects).
The initial input settings to achieve non-physiological gait patterns (scenarios and observations in Table~\ref{tb:experiment}) were chosen manually and purposefully by experienced therapists individually for each subject until the respective observation was present. 
The guidance force was set to 100\% for all trials, i.e., the subjects were walked by the Lokomat.
The treadmill speed was varied between 1.4km/h and 2.3km/h.

We conducted 63 experimental trials with the proposed algorithm in closed-loop with the ten nondisabled subjects, where each subject underwent at least five trials.
The difference in the number of experimental trials is due to each subject's availability. 
However, the test scenarios were chosen so that each scenario was tested comparably often.
Similarly to the data collection, the subjects were instructed to be as passive as possible.
Two therapists assessed the input adaptations suggested by the algorithm and rated whether the gait was physiological.
The therapists implemented the input adaptations until the algorithm indicated that a physiological gait cycle had been reached.
Additionally, the therapists indicated when they thought that a physiological gait had been reached and the algorithm should be stopped.

\subsection{Results}
\begin{figure}[t!]
      \centering
     \includegraphics[width=0.9975\columnwidth]{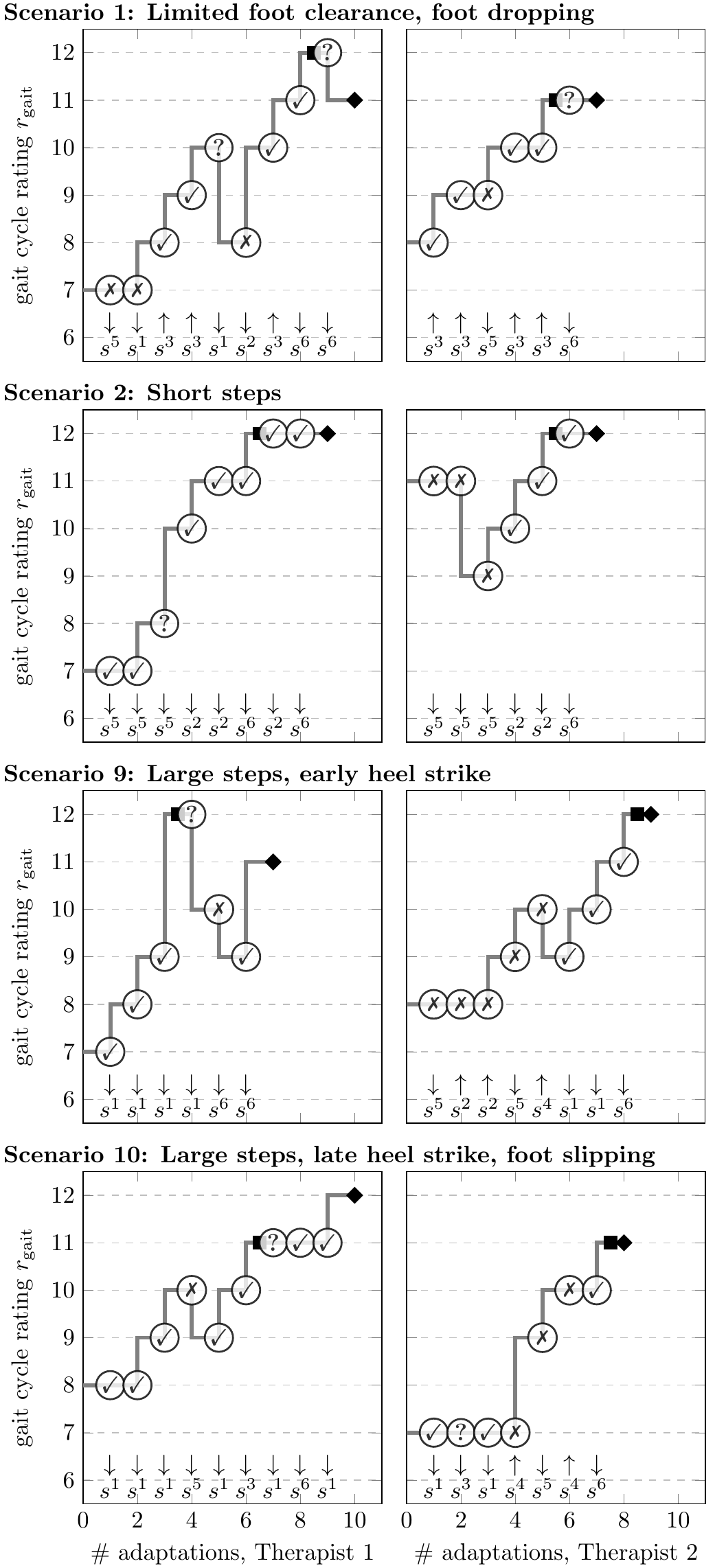}
      \caption{
      Experimental evaluation of the closed-loop recommendation system.
       Averaged for the eight experiments, a physiological gait was reached after 6.0 input adaptations (until square marker).
     The diamond marker indicates that the algorithm assessed the gait as physiological (stopping criterion).
      }
      \label{fig:experiment}
\end{figure}

Fig.~\ref{fig:experiment} illustrates eight representative trials with the first subject.
It contains four types of information and is separated by therapist in columns and by test scenario in rows:
\begin{enumerate}[i)]
\item The gait cycle rating $r_{\rm gait}$ ($y$-axis), calculated as the sum of the individual phase ratings  $r_{\rm gait}=r_{\rm HS}+r_{\rm MS}+r_{\rm TO}+r_{\rm SW}$, over the number of input adaptations ($x$-axis); 
\item the applied input adaptations and their direction, e.g., $s_1 \uparrow$ represents an increase of Setting~1 by one unit;
\item a statement from the therapists about the algorithm's suggested input adaptation, i.e., agreement with the suggestion as check mark $\checkmark$, disagreement as cross \ding{55}, and uncertainty about the suggestion as question mark \textbf{?}; and
\item the reaching of a physiological gait judged by the therapist with square markers {\tiny $ \blacksquare$} (for the usage as recommendation system) and by the algorithm with diamond markers {\small$\blacklozenge$} (for the usage as automatic adaptation system).
\end{enumerate}

In all eight illustrated experiments, the algorithm provides a reliable, although not monotonic, improvement in the physiology of the gait.
The input adaptations suggested by the algorithm led to a physiological gait for both the usage as recommendation system (square marker) and automatic adaptation system (diamond marker) in less than 10 adaptations with an overall rating of greater than or equal to 11, where 12 is the maximum possible rating.
The input adaptations during the test of Scenario~1 with both therapists (first row) are similar to the heuristic guidelines in Table~\ref{tb:experiment}, i.e., an increase in the knee range of motion ($s^3 \uparrow$).
The input adaptations for Scenario~2 (second row) are different from the heuristic guidelines.
Here, the algorithm converges to a kinematically different but physiological gait that is achieved through a slower treadmill speed and input settings that are adjusted accordingly.
For Scenario~9 and~10 (third and fourth row), the algorithm achieved a physiological gait through adaptations that are similar to the heuristic guidelines.
In all illustrated cases, the algorithm converges to a physiological gait.

Table~\ref{tb:experimentVal} summarizes all 63 experimental trials with ten subjects. 
On average, after a proposed input adaptation, the gait cycle improved in 63\% and did not degrade in 93\% of adaptations.
The latter percentage is important as sometimes, changing an input setting by only one unit is too small to make a noticeable change in the gait cycle and a couple of consecutive adaptations are necessary, e.g., for the orthosis speed ($s^6$).
Most importantly, a physiological gait cycle was reached for all trials within 6.0 \underline adaptations \underline per \underline trial (APT) on average for all subjects combined, and between 3.8 and 7.4 for each individual subject.
\begin{table}[t]
\setlength{\tabcolsep}{3.5pt}
\caption{
Summary of all Experimental Trials
}
\begin{center}
\def\arraystretch{1.1}
\label{tb:experimentVal}
\begin{tabular}{rlcccccc } 
\hline
  & &  &  &\multicolumn{2}{c}{Physiology of gait} \\
\multicolumn{2}{l}{Subject (body type)} & Trials & APT  
& improved & not degraded \\
\hline
1 & (193cm, 93kg, male) & 8  & 6.0 & 65\% & 92\%   
 \\
2 & (195cm, 100kg, male) & 5  & 3.8 & 89\% & 100\%   
 \\
3 & (163cm, 53kg, female)  & 6  & 6.8  & 68\% & 98\%  
 \\
4 & (175cm, 85kg, female) & 5  & 4.8 & 58\% & 92\%
 \\
5 & (172cm, 68kg, female)  & 9  & 4.9 & 57\% & 89\%  
 \\
6 & (190cm, 85kg, male)  & 5  & 7.4 & 54\% & 97\%  
 \\
7 & (167cm, 85kg, male)  & 6  & 7.2   & 60\% & 93\%  
 \\
8 & (180cm, 75kg, male)  &  5 & 7.0 & 60\% & 83\%
 \\
9 & (167cm, 64kg, female) & 7 &  5.7 & 65\% & 97\%
 \\
10 & (161cm, 48kg, female)  &  7 & 6.9 & 71\% & 92\%
 \\
\hline
\multicolumn{2}{l}{Overall} & 63 & 6.0 & 63\% & 93\% 
 \\
\hline
\end{tabular}
\end{center}
\end{table} 

\subsection{Discussion}
In general, the algorithm reached a physiological gait cycle within very few adaptations.
This is achieved as the proposed algorithm reasons about the gait cycle using the reward model in a higher dimensional feature space rather than the space of input settings.
As a result, the controller does not rely on trial and error search and, therefore, does not require to be tuned individually for each patient, which makes the approach more efficient, e.g., compared to classical reinforcement learning methods.
The majority of times, the therapists agreed with the suggestions from the algorithm, i.e., the suggested adaptations were conform with the heuristic tuning guidelines and their experience. 
Consequently, the resulting gait cycle was mostly kinematically similar to the one that the therapists would have chosen.
In some notable instances, the therapists disagreed or were uncertain about the proposition and were surprised by the improvement in the gait cycle, e.g., Row~1, Therapist~1, Adaptation~6; Row~2, Therapist~1, Adaptation~3; or Row~4, Therapist~2, Adaptation~4 in Fig.~\ref{fig:experiment}.
These instances are examples of situations where the algorithm chooses input adaptations, which were unknown to the therapists.
In these cases, the resulting gait cycle was sometimes kinematically different to the heuristic guidelines, e.g., a gait with slower treadmill speed.
Table~\ref{tb:experimentVal} shows that the algorithm is able to cope with various body types with similar results for all individuals.

It is worth noting that the differences between similar scenarios with two different therapists in Fig.~\ref{fig:experiment} and the same initial input settings do not necessarily lead to the same adjustment of input settings.
This observation can be explained as the physiology of the gait does not only depend on the chosen input settings but also on the hardware setup, e.g., the tightness of the straps, which differs slightly between therapists.
However, even though the hardware was setup slightly differently by the two therapists, the algorithm managed to find input settings that walk the subject physiologically, indicating a certain robustness to slight variations in the hardware.

\section{Conclusion and Future Work}
This paper has derived a supervised learning-based method utilizing human ratings for learning parameters of a feedback controller.
The approach was applied to the Lokomat robotic gait trainer with the goal of automatically adjusting the input settings to reach a physiological gait cycle by encapsulating the therapists' expertise in a reward model.
Feedback control was enabled by this reward model and a steady-state model, which allows for converting desired changes in gait into input adaptations.
Experiments with human subjects showed that the therapists' expertise in the form of ratings of four gait phases provides sufficient information to discriminate between physiological and non-physiological gait cycles. 
Furthermore, the provided adaptations led to an improvement of the gait cycle towards a physiological one within fewer than ten adaptations. 
The physiological gait cycle was partly reached by changes in input settings that domain experts would not have chosen themselves, suggesting that the proposed method might also be capable of generalizing from ratings and proposing improved settings for unseen scenarios. 
This observation remains to be confirmed with more data in future work.

Future work involves the data collection, evaluation, and validation of the proposed method with impaired patients.
This will include the assessment of asymmetric gait adaptations for the right and left legs, which can readily be achieved by considering one feature vector for each leg. 
Further, physical limitations and/or constraints in the patients' movements could be assessed online using sensor measurements of the Lokomat and considered for the selection of input settings.


\section*{Acknowledgment}
We gratefully acknowledge Patricia Andreia Gon\c calves Rodrigues, Ursula Costa, and Serena Maggioni for their clinical support and invaluable discussions; Luca Somaini and Nils Reinert for their technical support; and Liliana Pavel for implementing the user interface.


\addtolength{\textheight}{-1.75cm}

\bibliographystyle{IEEEtran}
\bibliography{root.bbl}

\begin{thebibliography}{10}
\providecommand{\url}[1]{#1}
\csname url@samestyle\endcsname
\providecommand{\newblock}{\relax}
\providecommand{\bibinfo}[2]{#2}
\providecommand{\BIBentrySTDinterwordspacing}{\spaceskip=0pt\relax}
\providecommand{\BIBentryALTinterwordstretchfactor}{4}
\providecommand{\BIBentryALTinterwordspacing}{\spaceskip=\fontdimen2\font plus
\BIBentryALTinterwordstretchfactor\fontdimen3\font minus
  \fontdimen4\font\relax}
\providecommand{\BIBforeignlanguage}[2]{{%
\expandafter\ifx\csname l@#1\endcsname\relax
\typeout{** WARNING: IEEEtran.bst: No hyphenation pattern has been}%
\typeout{** loaded for the language `#1'. Using the pattern for}%
\typeout{** the default language instead.}%
\else
\language=\csname l@#1\endcsname
\fi
#2}}
\providecommand{\BIBdecl}{\relax}
\BIBdecl

\bibitem{colombo2000}
G.~Colombo, M.~Joerg, R.~Schreier, and V.~Dietz, ``Treadmill training of
  paraplegic patients using a robotic orthosis,'' \emph{J. Rehab. Res. and
  Develop.}, vol.~37, no.~6, pp. 693--700, 2000.

\bibitem{reinkensmeyer2004robotics}
D.~J. Reinkensmeyer, J.~L. Emken, and S.~C. Cramer, ``Robotics, motor learning,
  and neurologic recovery,'' \emph{Annu. Rev. Biomed. Eng.}, vol.~6, pp.
  497--525, 2004.

\bibitem{emken2005robot}
J.~L. Emken and D.~J. Reinkensmeyer, ``Robot-enhanced motor learning:
  accelerating internal model formation during locomotion by transient dynamic
  amplification,'' \emph{{IEEE} Trans. Neural Syst. Rehab. Eng.}, vol.~13,
  no.~1, pp. 33--39, 2005.

\bibitem{marchal2009review}
L.~Marchal-Crespo and D.~J. Reinkensmeyer, ``Review of control strategies for
  robotic movement training after neurologic injury,'' \emph{J. Neuroeng. and
  Rehabil.}, vol.~6, no.~1, p.~20, 2009.

\bibitem{lambercy2012robots}
O.~Lambercy, L.~L{\"u}nenburger, R.~Gassert, and M.~Bolliger, ``Robots for
  measurement/clinical assessment,'' in \emph{Neurorehabil. Technol.}\hskip 1em
  plus 0.5em minus 0.4em\relax London: Springer, 2012, pp. 443--456.

\bibitem{van2016rehabilitation}
H.~M. Van~der Loos, D.~J. Reinkensmeyer, and E.~Guglielmelli, ``Rehabilitation
  and health care robotics,'' in \emph{Springer handbook of robotics}.\hskip
  1em plus 0.5em minus 0.4em\relax Springer, 2016, pp. 1685--1728.

\bibitem{HocomaNotes15}
Hocoma, ``Instructor script {Lokomat} training,'' Hocoma AG, Tech. Rep., 2015.

\bibitem{Colombo2004}
S.~Jezernik, G.~Colombo, and M.~Morari, ``Automatic gait-pattern adaptation
  algorithms for rehabilitation with a 4-dof robotic orthosis,'' \emph{{IEEE}
  J. Robot. Automat.}, vol.~20, no.~3, pp. 574--582, 2004.

\bibitem{Riener2005}
R.~Riener, L.~L{\"{u}}nenburger, S.~Jezernik, M.~Anderschitz, G.~Colombo, and
  V.~Dietz, ``Patient-cooperative strategies for robot-aided treadmill
  training: First experimental results,'' \emph{{IEEE} Trans. Neural Syst.
  Rehab. Eng.}, vol.~13, no.~3, pp. 380--394, 2005.

\bibitem{Riener2006}
R.~Riener, L.~L{\"{u}}nenburger, and G.~Colombo, ``Human-centered robotics
  applied to gait training and assessment,'' \emph{J. Rehab. Res. and
  Develop.}, vol.~43, no.~5, pp. 679--694, 2006.

\bibitem{Duschau2007}
A.~Duschau-Wicke, J.~{Von Zitzewitz}, R.~Banz, and R.~Riener, ``Iterative
  learning synchronization of robotic rehabilitation tasks,'' \emph{Proc.
  {IEEE} Int. Conf. Rehabil. Robot.}, pp. 335--340, 2007.

\bibitem{duschau2010}
A.~Duschau-Wicke, J.~von Zitzewitz, A.~Caprez, L.~L{\"{u}}nenburger, and
  R.~Riener, ``Path control: a method for patient-cooperative robot-aided gait
  rehabilitation,'' \emph{{IEEE} Trans. Neural Syst. Rehab. Eng.}, vol.~18,
  no.~1, pp. 38--48, 2010.

\bibitem{Maggioni2015}
S.~Maggioni, L.~L{\"{u}}nenburger, R.~Riener, and A.~Melendez-Calderon,
  ``Robot-aided assessment of walking function based on an adaptive
  algorithm,'' \emph{Proc. {IEEE} Int. Conf. Rehabil. Robot.}, pp. 804--809,
  2015.

\bibitem{emken2007motor}
J.~L. Emken, R.~Benitez, A.~Sideris, J.~E. Bobrow, and D.~J. Reinkensmeyer,
  ``Motor adaptation as a greedy optimization of error and effort,'' \emph{J.
  Neurophysiol.}, vol.~97, no.~6, pp. 3997--4006, 2007.

\bibitem{koenig2011psychological}
A.~Koenig, X.~Omlin, L.~Zimmerli, M.~Sapa, C.~Krewer, M.~Bolliger,
  F.~M{\"u}ller, and R.~Riener, ``Psychological state estimation from
  physiological recordings during robot-assisted gait rehabilitation.''
  \emph{J. Rehabil. Res. \& Develop.}, vol.~48, no.~4, pp. 367--386, 2011.

\bibitem{Begg2005}
R.~Begg and J.~Kamruzzaman, ``A machine learning approach for automated
  recognition of movement patterns using basic, kinetic and kinematic gait
  data,'' \emph{J. Biomechanics}, vol.~38, no.~3, pp. 401--408, 2005.

\bibitem{Wu2007}
J.~Wu, J.~Wang, and L.~Liu, ``Feature extraction via {KPCA} for classification
  of gait patterns,'' \emph{Human Movement Sci.}, vol.~26, no.~3, pp. 393--411,
  2007.

\bibitem{Yang2012}
M.~Yang, H.~Zheng, H.~Wang, S.~McClean, J.~Hall, and N.~Harris, ``A machine
  learning approach to assessing gait patterns for complex regional pain
  syndrome,'' \emph{Med. Eng. {\&} Physics}, vol.~34, no.~6, pp. 740--746,
  2012.

\bibitem{Tahafchi2017}
P.~Tahafchi, R.~Molina, J.~A. Roper, K.~Sowalsky, C.~J. Hass, A.~Gunduz, M.~S.
  Okun, and J.~W. Judy, ``Freezing-of-gait detection using temporal, spatial,
  and physiological features with a support-vector-machine classifier,''
  \emph{IEEE Int. Conf. Eng. in Medicine and Biol. Soc.}, pp. 2867--2870, 2017.

\bibitem{Sutton1998}
R.~S. Sutton and A.~G. Barto, \emph{Reinforcement learning: An
  introduction}.\hskip 1em plus 0.5em minus 0.4em\relax MIT press, 1998.

\bibitem{Knox2009}
W.~B. Knox and P.~Stone, ``Interactively shaping agents via human
  reinforcement: The {TAMER} framework,'' \emph{Proc. 5th Int. Conf. Knowledge
  Capture}, pp. 9--16, 2009.

\bibitem{Pilarski2011}
P.~M. Pilarski, M.~R. Dawson, T.~Degris, F.~Fahimi, J.~P. Carey, and R.~S.
  Sutton, ``Online human training of a myoelectric prosthesis controller via
  actor-critic reinforcement learning,'' \emph{Proc. {IEEE} Int. Conf. Rehabil.
  Robot.}, pp. 134--140, 2011.

\bibitem{Griffith2013}
S.~Griffith, K.~Subramanian, and J.~Scholz, ``Policy shaping: Integrating human
  feedback with reinforcement learning,'' \emph{Adv. Neural Inform. Process.
  Syst.}, pp. 2625--2633, 2013.

\bibitem{Christiano2017}
P.~F. Christiano, J.~Leike, T.~Brown, M.~Martic, S.~Legg, and D.~Amodei, ``Deep
  reinforcement learning from human preferences,'' in \emph{Adv. Neural Inform.
  Process. Syst.}, 2017, pp. 4302--4310.

\bibitem{wilde2018learning}
N.~Wilde, D.~Kuli{\'c}, and S.~L. Smith, ``Learning user preferences in robot
  motion planning through interaction,'' in \emph{Proc. {IEEE} Int. Conf.
  Robot. and Automat.}, 2018, pp. 619--626.

\bibitem{dorsa2017active}
A.~D.~D. Dorsa~Sadigh, S.~Sastry, and S.~A. Seshia, ``Active preference-based
  learning of reward functions,'' in \emph{Robot.: Sci. and Syst.}, 2017.

\bibitem{basu2018learning}
C.~Basu, M.~Singhal, and A.~D. Dragan, ``Learning from richer human guidance:
  Augmenting comparison-based learning with feature queries,'' in \emph{Proc.
  ACM/IEEE Int. Conf. Human-Robot Interaction}, 2018, pp. 132--140.

\bibitem{Menner2018}
M.~Menner and M.~N. Zeilinger, ``A user comfort model and index policy for
  personalizing discrete controller decisions,'' in \emph{Proc. Eur. Control
  Conf.}, 2018, pp. 1759--1765.

\bibitem{Kalman1964}
R.~E. Kalman, ``When is a linear control system optimal?'' \emph{J. Basic
  Eng.}, vol.~86, no.~1, pp. 51--60, 1964.

\bibitem{ng2000algorithms}
A.~Y. Ng and S.~J. Russell, ``Algorithms for inverse reinforcement learning.''
  in \emph{Proc. 17th Int. Conf. Mach. Learning}, 2000, pp. 663--670.

\bibitem{hewing2019learning}
L.~Hewing, K.~P. Wabersich, M.~Menner, and M.~N. Zeilinger, ``Learning-based
  model predictive control: Toward safe learning in control,'' \emph{Annu. Rev.
  Control, Robot., and Auton. Syst.}, vol.~3, 2020.

\bibitem{ravichandar2020recent}
H.~Ravichandar, A.~S. Polydoros, S.~Chernova, and A.~Billard, ``Recent advances
  in robot learning from demonstration,'' \emph{Annu. Rev. Control, Robot., and
  Auton. Syst.}, vol.~3, 2020.

\bibitem{Mombaur2010human}
K.~Mombaur, A.~Truong, and J.-P. Laumond, ``From human to humanoid locomotion:
  An inverse optimal control approach,'' \emph{Auton. Robots}, vol.~28, no.~3,
  pp. 369--383, 2010.

\bibitem{clever2016inverse}
D.~Clever, R.~M. Schemschat, M.~L. Felis, and K.~Mombaur, ``Inverse optimal
  control based identification of optimality criteria in whole-body human
  walking on level ground,'' in \emph{6th IEEE Int. Conf. Biomed. Robot. and
  Biomechatronics}, 2016, pp. 1192--1199.

\bibitem{kleesattel2018inverse}
A.~L. E.~N. Kleesattel and K.~Mombaur, ``Inverse optimal control based
  enhancement of sprinting motion analysis with and without running-specific
  prostheses,'' in \emph{7th IEEE Int. Conf. Biomed. Robot. and
  Biomechatronics}, 2018, pp. 556--562.

\bibitem{menner2018predictive}
M.~Menner, P.~Worsnop, and M.~N. Zeilinger, ``Constrained inverse optimal
  control with application to a human manipulation task,'' \emph{IEEE Trans.
  Control Syst. Technol.}, 2019, doi:10.1109/TCST.2019.2955663.

\bibitem{lin2016human}
J.~F.-S. Lin, V.~Bonnet, A.~M. Panchea, N.~Ramdani, G.~Venture, and
  D.~Kuli{\'c}, ``Human motion segmentation using cost weights recovered from
  inverse optimal control,'' in \emph{IEEE-RAS 16th Int, Conf. Humanoid
  Robots}, 2016, pp. 1107--1113.

\bibitem{Englert2017}
P.~Englert, N.~A. Vien, and M.~Toussaint, ``Inverse {KKT}: Learning cost
  functions of manipulation tasks from demonstrations,'' \emph{Int. J. Robot.
  Res.}, vol.~36, no. 13--14, pp. 1474--1488, 2017.

\bibitem{Menner2019cdc}
M.~Menner, K.~Berntorp, M.~N. Zeilinger, and S.~Di~Cairano, ``Inverse learning
  for human-adaptive motion planning,'' in \emph{Proc. 58th {IEEE} Conf.
  Decision and Control}, 2019, pp. 809--815.

\bibitem{Menner2018ecc}
M.~Menner and M.~N. Zeilinger, ``Convex formulations and algebraic solutions
  for linear quadratic inverse optimal control problems,'' in \emph{Proc. Eur.
  Control Conf.}, 2018, pp. 2107--2112.

\bibitem{abbeel2004apprenticeship}
P.~Abbeel and A.~Y. Ng, ``Apprenticeship learning via inverse reinforcement
  learning,'' in \emph{Proc. 21st Int. Conf. Mach. Learning}, 2004, p.~1.

\bibitem{ziebart2008maximum}
B.~D. Ziebart, A.~L. Maas, J.~A. Bagnell, and A.~K. Dey, ``Maximum entropy
  inverse reinforcement learning.'' in \emph{Proc. 23rd AAAI Conf. Artif.
  Intell.}, vol.~8, 2008, pp. 1433--1438.

\bibitem{levine2011nonlinear}
S.~Levine, Z.~Popovic, and V.~Koltun, ``Nonlinear inverse reinforcement
  learning with {Gaussian} processes,'' in \emph{Adv. Neural Inform. Process.
  Syst.}, 2011, pp. 19--27.

\bibitem{Hadfield2016}
D.~Hadfield-Menell, S.~J. Russell, P.~Abbeel, and A.~Dragan, ``Cooperative
  inverse reinforcement learning,'' in \emph{Adv. Neural Inform. Process.
  Syst.}, 2016, pp. 3909--3917.

\bibitem{bogert2016expectation}
K.~Bogert, J.~F.-S. Lin, P.~Doshi, and D.~Kuli{\'c}, ``Expectation-maximization
  for inverse reinforcement learning with hidden data,'' in \emph{Proc. Int.
  Conf. Auton. Agents \& Multiagent Syst.}, 2016, pp. 1034--1042.

\bibitem{joukov2017gaussian}
V.~Joukov and D.~Kuli{\'c}, ``Gaussian process based model predictive
  controller for imitation learning,'' in \emph{IEEE-RAS 17th Int. Conf.
  Humanoid Robot.}, 2017, pp. 850--855.

\bibitem{Bishop2006}
C.~M. Bishop, \emph{Pattern Recognition and Machine Learning}.\hskip 1em plus
  0.5em minus 0.4em\relax Secaucus, NJ: Springer, 2006.

\bibitem{riener2010locomotor}
R.~Riener, L.~L{\"u}nenburger, I.~C. Maier, G.~Colombo, and V.~Dietz,
  ``Locomotor training in subjects with sensori-motor deficits: an overview of
  the robotic gait orthosis lokomat,'' \emph{J. Healthcare Eng.}, vol.~1,
  no.~2, pp. 197--216, 2010.

\bibitem{Perry1992}
J.~Perry, \emph{Gait Analysis: Normal and Pathological Function}.\hskip 1em
  plus 0.5em minus 0.4em\relax Thorofare, NJ: SLACK Incorporated Inc., 1992.

\bibitem{James2013}
G.~James, D.~Witten, T.~Hastie, and R.~Tibshirani, \emph{An introduction to
  statistical learning}.\hskip 1em plus 0.5em minus 0.4em\relax New York:
  Springer, 2013, vol. 112.

\end{thebibliography}

\end{document}